\documentclass[10pt]{article}
\usepackage[utf8]{inputenc} 
\usepackage[T1]{fontenc}    
\usepackage{lmodern}
\usepackage{hyperref}       
\usepackage{url}            
\usepackage{booktabs}       
\usepackage{amsfonts}       
\usepackage{microtype}      
\usepackage{algorithm}
\usepackage{algorithmic}
\usepackage{amsmath, amsthm, amssymb, multirow}
\usepackage{graphicx}
\usepackage{wrapfig}
\usepackage[body={6.5in,8in}]{geometry}
\usepackage{epsfig}
\usepackage{xcolor}
\usepackage{soul}
\usepackage{mathrsfs}
\usepackage{subfigure}
\usepackage{array}
\usepackage{epstopdf}
\usepackage{amsbsy,bm,theorem,ifthen,color}

\title{Incorporating Multiple Cluster Centers for Multi-Label Learning}

\author{
  Senlin Shu$^1$, Fengmao Lv$^{2}$, Yan Yan$^{3}$, Li Li$^{1}$, Shuo He$^{4}$, Jun He$^{1}$\\
  $^1$Southwest University, Chongqing, China\\
  $^2$Southwest Jiaotong University, Chengdu, China\\
  $^3$Northwestern Polytechnical University, Xian, China \\
  $^4$University of Electronic Science and Technology, Chengdu, China\\
}
\begin{document}
\maketitle





\begin{abstract}
	Multi-label learning deals with the problem that each instance is associated with multiple labels simultaneously. Most of the existing approaches aim to improve the performance of multi-label learning by exploiting label correlations. Although the data augmentation technique is widely used in many machine learning tasks, it is still unclear whether data augmentation is helpful to multi-label learning. In this article, we propose to leverage the data augmentation technique to improve the performance of multi-label learning. Specifically, we first propose a novel data augmentation approach that performs clustering on the real examples and treats the cluster centers as virtual examples, and these virtual examples naturally embody the local label correlations and label importances. Then, motivated by the cluster assumption that examples in the same cluster should have the same label, we propose a novel regularization term to bridge the gap between the real examples and virtual examples, which can promote the local smoothness of the learning function. Extensive experimental results on a number of real-world multi-label datasets clearly demonstrate that our proposed approach outperforms the state-of-the-art counterparts.
\end{abstract}


\section{Introduction}
Multi-label learning deals with the problem that each instance is associated with multiple labels simultaneously. Due to its ability to cope with the real-world objects with multiple semantic meanings, multi-label learning has been successfully applied in various application domains~\cite{zhang2014review}, such as tag recommendation~\cite{song2008sparse}, bioinformatics~\cite{elisseeff2002kernel,zhang2006multilabel}, information retrieval~\cite{gopal2010multilabel,zhu2005multi}, rule mining~\cite{thabtah2004mmac,veloso2007multi}, web mining~\cite{kazawa2005maximal,tang2009large}, and so on.
Formally speaking, suppose the given multi-label data set is denoted by $\mathcal{D}=\{\mathbf{x}_i,\mathbf{y}_i\}_{i=1}^n$ where $\mathbf{x}_i\in\mathbb{R}^d$ is a feature vector with $d$ dimensions (features) and $\mathbf{y}_i\in\{-1,+1\}^q$ is the corresponding label vector with the size of label space being $q$. Here, $y_{ij}=1$ indicates that the $i$-th instance $\mathbf{x}_i$ has the $j$-th label (or equivalently, the $j$-th label is a relevant label of $\mathbf{x}_i$), otherwise the $j$-th label is an irrelevant label of $\mathbf{x}_i$. Let $\mathcal{X}=\mathbb{R}^d$ be the $d$-dimensional feature space, and $\mathcal{Y}=\{-1,+1\}^q$ be the $q$-dimensional label space, multi-label learning aims to induce a mapping function $f:\mathcal{X}\rightarrow\mathcal{Y}$, which is able to correctly predict the label vector of unseen instances.

To solve the multi-label learning problem, the most straightforward solution is Binary Relevance (BR)~\cite{boutell2004learning,tsoumakas2009mining}, which aims to decompose the original learning problem into a set of independent binary classification problems. However, this solution generally achieves mediocre performance, as label correlations are regrettably ignored. To solve this problem, a large number of multi-label learning approaches take into account label correlations explicitly or implicitly to improve the learning performance, such as chains of binary classification~\cite{read2011classifier}, ensemble of multi-class classification~\cite{tsoumakas2011random}, label-specific features~\cite{zhang2014lift,huang2016learning}, and feature selection~\cite{huang2018joint}.

Although a considerable number of methods have been proposed to improve the performance of multi-label learning, it still remains unknown whether data augmentation is helpful to multi-label learning. Data augmentation~\cite{inoue2018data,zhang2017mixup,perez2017effectiveness} is a widely used technique in many machine learning tasks, and it aims to apply a small mutation in the original training data and synthetically creating new examples to virtually increase the number of training examples, thereby achieving better generalization performance. In this article, we propose to leverage the data augmentation technique to improve the performance of multi-label learning. We show that the data augmentation technique can not only capture the local label correlations and label importances, but also potentially enables the learning function to be smooth. Specifically, our proposed data augmentation method is motivated by the statement that the local data characteristics can be captured by clustering~\cite{jain1988algorithms,jain1999data}. The cluster center is an average feature vector of all the instances in the cluster, which can also be regarded as a local representative of the cluster.
If we consider the cluster center as a new instance, its corresponding label vector (labeling information) is supposed to the average label vector of all the instances in the cluster. Such data augmentation approach brings multiple important advantages for multi-label learning. First,
the local label correlations (in the cluster) can be captured by the label vector of the cluster center. The local label correlations are also already shown to be very helpful to multi-label learning by existing works~\cite{huang2012multi1,zhu2018multi}.
Second, the labeling importance degree of each label in the cluster can be reflected by the label vector of the cluster center. Many existing multi-label learning approaches~\cite{li2015leveraging,hou2016multi,zhang2018feature,he2018estimating} have shown that great performance can be achieved by taking into account the labeling importance degree of each relevant label. Third, each cluster center can also be considered as the \textsl{label smoothing}~\cite{pereyra2017regularizing} of all the instances in the cluster.
Note that the label vector of each real instance is binary ($\{-1,+1\}^q$), while the label vector of the cluster center is continuous ($[-1,+1]^q$), which potentially makes the learning function smoother. In addition, our proposed augmentation approach can be considered as a generalization of the popular $\textsl{mixup}$ approach~\cite{zhang2017mixup} to the case of multiple examples.

With the augmented training data at hand, we further propose a novel regularization term. Inspired by the $\textsl{cluster assumption}$~\cite{chapelle2003cluster,zhou2004learning} that instances in the same cluster are supposed to have the same label, we present a novel regularization term to bridge the gap between the real examples and the virtual examples. Specifically, the modeling output of each real instance and its corresponding cluster center should be similar. Such a regularization term naturally promotes the local smoothness of the learning function. The effectiveness of the proposed approach is clearly demonstrated by extensive experimental results on a number of real-world multi-label datasets.

In summary, our main contributions are three-fold:
\begin{itemize}
	\item We propose a novel data augmentation approach to enlarge the multi-label training set by generating multiple compact examples.
	\item We propose a novel regularization approach that bridges the gap between the real examples and the virtual examples, by encouraging the modeling output of each real example to be similar to that of the corresponding virtual example.
	\item In order to perform nonlinear classification, we extend our model to a kernel-based nonlinear model. For optimizing the final objective function, we derive globally optimal solutions (e.g., closed-form solutions).
\end{itemize}

Extensive experimental results clearly demonstrate that our proposed approach outperforms the state-of-the-art counterparts.

The rest of the paper is organized as follows. Section 2 briefly reviews existing approaches for multi-label learning and data augmentation. Section 3 presents the technical details of our proposed approach. Section 4 reports the experimental results of comparative studies. Section 5 concludes this article.

\section{Related Work}

\subsection{Multi-Label Learning} 
Multi-label learning~\cite{liu2015optimality,liu2015large,chen2019multi,chen2019two,shen2018deep,huang2019supervised,feng2020regularized} deals with the problem that each instance is associated with multiple labels simultaneously. So far, a huge number of approaches have been proposed to deal with the multi-label learning problem. According to the \emph{order of label correlations}, most of the existing approaches could be roughly divided into three categories. Approaches in the first category~\cite{boutell2004learning,zhang2007ml} do not take label correlations into consideration, and normally tackle the multi-label learning problem in a label-by-label manner, such as binary relevance~\cite{boutell2004learning,tsoumakas2009mining}, algorithm adaption approaches~\cite{zhang2007ml} and multi-label learning with label-specific features~\cite{zhang2014lift,huang2016learning}. Although this kind of approaches is simple and intuitive, it can only achieve passable performance, due to the neglect of label correlations. To address this drawback, approaches in the second category~\cite{elisseeff2002kernel,furnkranz2008multilabel} take into account the pairwise (second-order) correlations between labels.
One way to consider pairwise relationships is to exploit the
interactions between pairs of labels, such as calibrated label
ranking, multi-label teaching-to-learn
and learning-to-teach~\cite{gongt2016teaching}, and joint feature selection and classification~\cite{huang2018joint}. 
Another way is to impose the ranking criterion, which can be
incorporated into the objective function to be optimized
by learning models such as RankSVM~\cite{elisseeff2002kernel}, maximum entropy classifiers~\cite{furnkranz2008multilabel}, and relative labeling-importance aware multi-label
learning~\cite{li2015leveraging}. In addition, approaches in the third category~\cite{read2011classifier,tsoumakas2011random} consider high-order correlations among multiple labels. One way is to model interactions among all class labels, i.e., to consider the influences of all other labels on each label, such as linear combination~\cite{cheng2009combining}, and collaboration based multi-label~\cite{feng2019collaboration}. Another way is to model interactions among a subset of class labels instead of all of them, such as classifier chains~\cite{read2011classifier}, and RAkEL~\cite{tsoumakas2011random}. Note that the existing approaches only exploit label correlations from the given training examples, and there still remains the question of whether we can exploit label correlations from virtual examples.

\subsection{Data Augmentation}
This is a widely used technique in many machine learning tasks, and it aims to apply a small mutation on the original training data and synthetically creating new examples to virtually increase the number of training examples. Traditional data augmentation techniques~\cite{inoue2018data} for image classification tasks normally generate new examples from the original training data by flipping, distorting, adding a small amount of noise, or cropping a patch from an original image. Apart from the traditional data augmentation techniques, the \textsl{SimplePairing} approach~\cite{inoue2018data} randomly chooses two examples $(\mathbf{x}_a,\mathbf{y}_a)$ and $(\mathbf{x}_b,\mathbf{y}_b)$, then the new example is generated (randomly decided) by either $(\frac{\mathbf{x}_a+\mathbf{x}_b}{2},\mathbf{y}_a)$ or $(\frac{\mathbf{x}_a+\mathbf{x}_b}{2},\mathbf{y}_b)$. On the other hand, given such two examples, the new example generated by the \textsl{mixup} approach~\cite{zhang2017mixup} is represented as $(\frac{\mathbf{x}_a+\mathbf{x}_b}{2},\frac{\mathbf{y}_a+\mathbf{y}_b}{2})$. Although satisfied performance has been achieved by the two approaches, they only focus on generating new examples by manipulating exactly two real examples. How to generate new examples from multiple real examples and how to apply the generated new examples for improving the performance of multi-label learning task still remains unknown. These questions will be answered in the next section.
\section{The Proposed Approach}
In this section, we present our approach IMCC (Incorporating Multiple Clustering Centers). Following the notations used in Introduction, we denote the feature matrix by $\mathbf{X}=[\mathbf{x}_1,\mathbf{x}_2,\cdots,\mathbf{x}_n]^\top\in\mathbb{R}^{n\times d}$ and denote the label matrix by $\mathbf{Y}=[\mathbf{y}_1,\mathbf{y}_2,\cdots,\mathbf{y}_n]^\top\in\{-1,+1\}^{n\times q}$, where $n$ is the number of examples. IMCC works by taking two elementary steps, including virtual examples generation and multi-label model training.

Before presenting the details of our proposed approach, we would like to introduce the following three methodological phases of our approach: 1) We first use $k$-means method to cluster real examples and take the cluster centers as generated virtual examples.
2) We propose a novel regularization term to bridge the gap between the real examples and virtual examples while training the desired model. 3) In order to perform nonlinear classification, we extend our model to a kernel-based nonlinear model.
\subsection{Virtual Examples Generation}
In the first step, IMCC aims to generate a number of virtual examples that could be useful to the subsequent model training step. In order to generate new examples, we have to gain some insights from the existing examples. To achieve this, the clustering techniques are widely used as stand-alone tools for data analysis~\cite{zhang2014lift}. In the paper, the popular $k$-means algorithm~\cite{jain1999data} is adopted, due to its simplicity and effectiveness. Suppose the instances are partitioned into $c$ disjoint clusters $\{\mathcal{Z}_1,\mathcal{Z}_2,\cdots,\mathcal{Z}_c\}$. If the $i$-th instance $\mathbf{x}_i$ is partitioned into the $j$-th cluster $\mathcal{Z}_j$, then $\mathbf{x}_i\in\mathcal{Z}_j$. Typically, the clustering center is a representative instance of the cluster, hence its semantic meanings could be the average of semantic meanings of all the instances in the cluster.
Hence for each cluster $\mathcal{Z}_i$, its clustering center $\mathbf{z}_j$ is defined as:
\begin{gather}
	\label{eq1}
	\mathbf{z}_j=\frac{1}{|\mathcal{Z}_j|}\sum_{i=1}^n\mathbf{x}_i\cdot\mathbb{I}(\mathbf{x}_i\in\mathcal{Z}_j),
\end{gather}
where $\mathbb{I}(\cdot)$ is a indicator function, i.e., $\mathbb{I}(\mathbf{x}_i\in\mathcal{Z}_j)$ equals 1 if $\mathbf{x}_i\in\mathcal{Z}_j$ is true, otherwise it equals 0. From one specific view of point, $\mathbf{z}_j$ is the local representative instance of the instances belonging to the $j$-th cluster, hence its semantic meanings could be the average of semantic meanings of all the instances in the cluster. In other words, suppose $\mathbf{t}_j$ denote the labeling information of $\mathbf{z}_j$, then $\mathbf{t}_j$ should be the average label vectors of all the instances in $\mathcal{Z}_j$:
\begin{gather}
	\label{eq2}
	\mathbf{t}_j=\frac{1}{|\mathcal{Z}_j|}\sum_{i=1}^n\mathbf{y}_i\cdot\mathbb{I}(\mathbf{x}_i\in\mathcal{Z}_j).
\end{gather}
In this way, we can have a complementary training set $\mathcal{D}^\prime=\{\mathbf{z}_j,\mathbf{t}_j\}_{j=1}^c$, where $c$ is a hyper-parameter that denotes the number of the clusters and we will empirically analyze the influence of $c$ in section 4.
Here we give a concrete example to illustrate the advantage of the proposed data augmentation approach. Suppose there is a cluster including three examples $(\mathbf{x}_a,\mathbf{y}_a)$, $(\mathbf{x}_b,\mathbf{y}_b)$ and $(\mathbf{x}_c,\mathbf{y}_c)$, where $\mathbf{y}_a=[1,-1,1,1]^\top$, $\mathbf{y}_b = [1,-1,-1,1]^\top$, $\mathbf{y}_c = [1, 1, -1, 1]^\top$. Hence the virtual example is given as $(\frac{\mathbf{x}_a+\mathbf{x}_b+\mathbf{x}_c}{3},\frac{\mathbf{y}_a+\mathbf{y}_b+\mathbf{y}_c}{3})$, where the label vector is $\frac{\mathbf{y}_a+\mathbf{y}_b+\mathbf{y}_c}{3}=[1,-\frac{2}{3},\frac{1}{3},1]^\top$. First, it is clearly that our proposed data augmentation approach could be considered as a generalization of the popular \textsl{mixup} approach~\cite{zhang2017mixup} to the case of multiple examples. Second, the generated label vector contains soft labels, which are able to describe the labeling importance degree of each label~\cite{hou2016multi,feng2018leveraging,zhang2018feature,feng2019partial} in the cluster. As we can see, the first and the fourth label are most important. Third, as each soft label vector is generated by aggregating the local labeling information in the cluster, the local label correlations could be captured. Concretely, it is clear that the first and the fourth label co-occur in the same cluster, hence they have very strong local correlations. Besides, there is a negative value for the second label, which suggests that the second label may possess the opposite semantic meaning against other labels, since other labels have a positive value. Fourth, the soft label vector of cluster center can also be considered as the \textsl{label smoothing}~\cite{pereyra2017regularizing} of all the instances in the cluster.
Note that the label vector of each real instance is binary ($\{-1,+1\}^q$), while the label vector of the cluster center is continuous ($[-1,+1]^q$), which potentially makes the learning function smoother.

\subsection{Multi-Label Model Training}
For compact representations of the complementary training set, the additional feature matrix and the corresponding label matrix are denoted by $\mathbf{Z}=[\mathbf{z}_1,\mathbf{z}_2,\cdots,\mathbf{z}_c]^\top\in\mathbb{R}^{c\times d}$ and $\mathbf{T}=[\mathbf{t}_1,\mathbf{t}_2,\cdots,\mathbf{t}_c]^\top\in[-1,+1]^{c\times q}$, respectively. Note that there are soft labels (ranging from -1 to +1) in $\mathbf{T}$ while hard labels (either -1 or +1 ) in $\mathbf{Y}$. 

With the original data set $\mathcal{D}$ and the complementary data set $\mathcal{D}^\prime$, the objective function could be designed as follows:
\begin{align}
	\label{eq3}
	\min_{\mathbf{W},\mathbf{b}}\frac{1}{2}\sum_{i=1}^n\left\|\mathbf{W}^{\top}\mathbf{x}_i+\mathbf{b}-\mathbf{y}_i\right\|_2^2 + \frac{\alpha}{2}\sum_{j=1}^c\left\|\mathbf{W}^{\top}\mathbf{z}_j+\mathbf{b}-\mathbf{t}_j\right\|_2^2
	+\frac{\beta}{2}\left\|\mathbf{W}\right\|_F^2,
\end{align}
where $\mathbf{W}\in\mathbb{R}^{d\times q}$ and $\mathbf{b}\in\mathbb{R}^q$ are the model parameters, and the widely used Frobenius norm of $\mathbf{W}$ is employed to reduce the model complexity to avoid overfitting. The trade-off hyperparameters $\alpha$ and $\beta$ control the importance of learning from virtual examples and model complexity, respectively. By a compact representation, problem~(\ref{eq3}) can be equivalently stated as follows:
\begin{align}
	\label{eq4}
	\min_{\mathbf{W},\mathbf{b}}\frac{1}{2}\left\|\mathbf{XW}+\mathbf{1}_n\mathbf{b}^\top-\mathbf{Y}\right\|_F^2 + \frac{\alpha}{2}\left\|\mathbf{ZW}+\mathbf{1}_c\mathbf{b}^\top-\mathbf{T}\right\|_F^2
	+\frac{\beta}{2}\left\|\mathbf{W}\right\|_F^2,
\end{align}
where $\mathbf{1}_n$ and $\mathbf{1}_c$ denote the vectors of size $n$ and $c$, with every element equals 1. Although the real examples and virtual examples models have been established in problem~(\ref{eq4}), there is still a gap between the real examples and virtual examples. Inspired by the \textsl{cluster assumption}~\cite{chapelle2003cluster,zhou2004learning} that instances in the same cluster are supposed to have the same label, we propose a novel regularization approach that the modeling output of each instance should be similar as that of the corresponding cluster center. Thus the regularization term is stated as:
\begin{gather}
	\label{eq5}
	\min_{\mathbf{W},\mathbf{b}}\sum_{i=1}^n\left\|(\mathbf{W}^{\top}\mathbf{x}_i+\mathbf{b})-(\mathbf{W}^{\top}\hat{\mathbf{z}}_i+\mathbf{b})\right\|_2^2,
\end{gather}
where $\widehat{\mathbf{z}}_i = \sum_{j=1}^c\mathbf{z}_j\cdot\mathbb{I}(\mathbf{x}_i\in\mathcal{Z}_j)$ denotes the center of the cluster, which $\mathbf{x}_{i}$ belongs to. Note that the clusters are disjoint, hence $\sum_{j=1}^c\mathbf{z}_j\cdot\mathbb{I}(\mathbf{x}_i\in\mathcal{Z}_j)$ results in only one cluster center $\mathbf{z}_j$ such that $\mathbf{x}_i\in\mathcal{Z}_j$ is true.
In this way, we can bridge the gap between the real examples and virtual examples, which can promote the local smoothness of the learning function.  Here, we specially introduce a matrix $\widehat{\mathbf{Z}} = [\widehat{\mathbf{z}}_1,\widehat{\mathbf{z}}_2,\cdots,\widehat{\mathbf{z}}_n]^\top\in\mathbb{R}^{n\times d}$.
In this way, problem~(\ref{eq5}) is equivalent to:
\begin{align}
	\label{eq6}
	\min_{\mathbf{W},\mathbf{b}}\left\|(\mathbf{XW}+\mathbf{1}_n\mathbf{b}^\top)-(\widehat{\mathbf{Z}}\mathbf{W}+\mathbf{1}_n\mathbf{b}^\top)\right\|_F^2
	=\min_{\mathbf{W}}\left\|\mathbf{XW}-\widehat{\mathbf{Z}}\mathbf{W}\right\|_F^2.
\end{align}
By combining problem~(\ref{eq4}) and problem~(\ref{eq6}), the final objective function is given as:
\begin{align}
	\label{eq7}
	\min_{\mathbf{W},\mathbf{b}}\frac{1}{2}\left\|\mathbf{XW}+\mathbf{1}_n\mathbf{b}^\top-\mathbf{Y}\right\|_F^2  + \frac{\alpha}{2}\left\|\mathbf{ZW}+\mathbf{1}_c\mathbf{b}^\top-\mathbf{T}\right\|_F^2
 +\frac{\beta}{2}\left\|\mathbf{W}
	\right\|_F^2 + \frac{\gamma}{2}\left\|\mathbf{XW}-\widehat{\mathbf{Z}}\mathbf{W}\right\|_F^2,
\end{align}
where $\gamma$ is a trade-off parameter that controls the importance of the regularization term.

\subsection{Optimization}

For optimization, it would be not hard to compute the derivative of problem~(\ref{eq7}) with respect to $\mathbf{W}$ and $\mathbf{b}$:
\begin{align}
	\label{eq8}
	\nabla_{\mathbf{W}} &=\Big(\mathbf{X}^\top\mathbf{X}+\alpha\mathbf{Z}^\top\mathbf{Z}+\beta\mathbf{I}+\gamma(\mathbf{X}-\widehat{\mathbf{Z}})^\top(\mathbf{X}-\widehat{\mathbf{Z}})\Big)\mathbf{W}+\mathbf{X}^\top(\mathbf{1}_n\mathbf{b}^\top-\mathbf{Y})+\alpha\mathbf{Z}^\top(\mathbf{1}_c\mathbf{b}^\top-\mathbf{T}),\\
	\nabla_{\mathbf{b}} &= (\mathbf{W}^\top\mathbf{X}^\top+\mathbf{b}\mathbf{1}_n^\top-\mathbf{Y}^\top)\mathbf{1}_n
	+\alpha(\mathbf{W}^\top\mathbf{Z}^\top+\mathbf{b}\mathbf{1}_c^\top-\mathbf{T}^\top)\mathbf{1}_c.
\end{align}
From the expression of the derivative, we can easily find that problem~(\ref{eq7}) has the closed-form solutions, which are the globally optimal solutions.
By setting $\nabla_{\mathbf{W}}$ and $\nabla_{\mathbf{b}}$ to 0, we can obtain:
\begin{align}
	\label{eq10}
	&\Big(\mathbf{X}^\top\mathbf{X}+\alpha\mathbf{Z}^\top\mathbf{Z}+\beta\mathbf{I}+\gamma(\mathbf{X}-\widehat{\mathbf{Z}})^\top(\mathbf{X}-\widehat{\mathbf{Z}})\Big)\mathbf{W} = -\mathbf{X}^\top(\mathbf{1}_n\mathbf{b}^\top-\mathbf{Y})-\alpha\mathbf{Z}^\top(\mathbf{1}_c\mathbf{b}^\top-\mathbf{T}), \\
	\label{eq11}
	&(n+ \alpha c)\mathbf{b} = (\mathbf{Y}^\top-\mathbf{W}^\top\mathbf{X}^\top)\mathbf{1}_n+\alpha(\mathbf{T}^\top-\mathbf{W}^\top\mathbf{Z}^\top)\mathbf{1}_c.
\end{align}
By substituting Eq. (\ref{eq11}) into Eq. (\ref{eq10}), we can obtain:
\begin{align}
	\nonumber
	&\Big(\mathbf{X}^\top\mathbf{X}+\alpha\mathbf{Z}^\top\mathbf{Z}+\beta\mathbf{I}+\gamma(\mathbf{X}-\widehat{\mathbf{Z}})^\top(\mathbf{X}-\widehat{\mathbf{Z}})\Big)\mathbf{W} = \\
	\nonumber
	&\quad\quad\quad\quad\quad\quad\quad\quad\quad\quad\quad\quad\quad\quad-\mathbf{X}^\top(\frac{1}{n+\alpha c}\mathbf{1}_n\Big((\mathbf{Y}^\top-\mathbf{W}^\top\mathbf{X}^\top)\mathbf{1}_n
	+\alpha(\mathbf{T}^\top
	-\mathbf{W}^\top\mathbf{Z}^\top)\mathbf{1}_c\Big)^\top-\mathbf{Y})\\
	\label{eq12}
	&\quad\quad\quad\quad\quad\quad\quad\quad\quad\quad\quad\quad\quad\quad-\alpha\mathbf{Z}^\top(\frac{1}{n+\alpha c}\mathbf{1}_c\Big((\mathbf{Y}^\top-\mathbf{W}^\top\mathbf{X}^\top)\mathbf{1}_n
	+\alpha(\mathbf{T}^\top
	-\mathbf{W}^\top\mathbf{Z}^\top)\mathbf{1}_c\Big)^\top-\mathbf{T}). 
\end{align}
Then we can rewrite the Eq. (\ref{eq12}) as:
\begin{align}
	\nonumber
	&\Big(\mathbf{X}^\top\mathbf{X}+\alpha\mathbf{Z}^\top\mathbf{Z}+\beta\mathbf{I}+\gamma(\mathbf{X}-\widehat{\mathbf{Z}})^\top(\mathbf{X}-\widehat{\mathbf{Z}})- \frac{\mathbf{X}^\top\mathbf{1}_n+\alpha\mathbf{Z}^\top\mathbf{1}_c}{n+\alpha c}
	(\mathbf{1}_n^\top\mathbf{X}+\alpha\mathbf{1}_c^\top\mathbf{Z})\Big)\mathbf{W} =\\
	\label{eq13}
	&\quad\quad\quad\quad\quad\quad\quad\quad\quad\quad\quad\quad\quad\quad\quad\quad\quad\quad\quad\quad\quad \Big(\mathbf{X}^\top\mathbf{Y}
	+\alpha\mathbf{Z}^\top\mathbf{T}-\frac{\mathbf{X}^\top\mathbf{1}_n+\alpha\mathbf{Z}^\top\mathbf{1}_c}{n+\alpha c}(\mathbf{1}_n^\top\mathbf{Y}+\alpha\mathbf{1}_c^\top\mathbf{T})\Big).
\end{align}
By solving the problem (\ref{eq13}), we can directly obtain the optimal $\mathbf{W}$ and the optimal values of $\mathbf{W}$ and $\mathbf{b}$ are shown as follows: 
\begin{align}
	\nonumber
	\mathbf{W} &= \Big(\mathbf{X}^\top\mathbf{X}+\alpha\mathbf{Z}^\top\mathbf{Z}+\beta\mathbf{I}+\gamma(\mathbf{X}-\widehat{\mathbf{Z}})^\top(\mathbf{X}-\widehat{\mathbf{Z}})
	- \frac{\mathbf{X}^\top\mathbf{1}_n+\alpha\mathbf{Z}^\top\mathbf{1}_c}{n+\alpha c}
	(\mathbf{1}_n^\top\mathbf{X}\\
	\label{eq14}
	&\quad\quad\quad\quad\quad\quad\quad\quad\quad\quad\quad\quad\quad+\alpha\mathbf{1}_c^\top\mathbf{Z})\Big)^{-1}\Big(\mathbf{X}^\top\mathbf{Y}
	+\alpha\mathbf{Z}^\top\mathbf{T}-\frac{\mathbf{X}^\top\mathbf{1}_n+\alpha\mathbf{Z}^\top\mathbf{1}_c}{n+\alpha c}(\mathbf{1}_n^\top\mathbf{Y}+\alpha\mathbf{1}_c^\top\mathbf{T})\Big),\\
	\label{eq15}
	\mathbf{b} &= \frac{1}{n+\alpha c}\Big((\mathbf{Y}^\top-\mathbf{W}^\top\mathbf{X}^\top)\mathbf{1}_n
	+\alpha(\mathbf{T}^\top
	-\mathbf{W}^\top\mathbf{Z}^\top)\mathbf{1}_c\Big).
\end{align}

\subsection{Kernel Extension}
In the previous section, we provided the closed-form solutions of the linear model. However, such simple linear model cannot work in the nonlinear case, which may deteriorate the learning performance when the data cannot be linearly separated. To address this problem, in this section, we show that our approach can be easily extended to a kernel-based nonlinear model. 

Specifically, we use a nonlinear feature mapping $\phi(\cdot):\mathbb{R}^d\rightarrow\mathbb{R}^{\mathcal{H}}$, which maps the original feature space to some higher (maybe infinite) dimensional Hilbert space, i.e., $\mathbf{X}\rightarrow\phi(\mathbf{X})$. By representation theorem~\cite{scholkopf2002learning}, the optimal value of $\mathbf{W}$ can be represented by a linear combination of the input features $\phi(\mathbf{X})$, which means $\mathbf{W}=\phi(\mathbf{X})^\top\mathbf{A}$ where $\mathbf{A}\in\mathbb{R}^{n\times q}$ is a coefficients matrix. In other words, $\mathbf{A}$ is a new variable that can be used to replace $\mathbf{W}$. Note that the kernel matrix is normally given as $\mathbf{K}=\phi(\mathbf{X})\phi(\mathbf{X})^\top\in\mathbb{R}^{n\times n}$, hence $\phi(\mathbf{X})\mathbf{W} = \phi(\mathbf{X})\phi(\mathbf{X})^\top\mathbf{A} = \mathbf{KA}$, where the element $k_{ij}$ of $\mathbf{K}$ is defined as $k_{ij}=\phi(\mathbf{x}_i)^\top\phi(\mathbf{x}_j)=\kappa(\mathbf{x}_i,\mathbf{x}_j)$, and $\kappa(\cdot,\cdot)$ denotes the kernel function. Similarly, $\phi(\mathbf{Z})\mathbf{W} = \phi(\mathbf{Z})\phi(\mathbf{X})^\top\mathbf{A} = \widetilde{\mathbf{K}}\mathbf{A}$ where $\widetilde{\mathbf{K}}\in\mathbb{R}^{c\times n}$ with its element $\widetilde{k}_{ij} = \phi(\mathbf{z}_i)^\top\phi(\mathbf{x}_j) = \kappa(\mathbf{z}_i,\mathbf{x}_j)$. In addition, $\phi(\widehat{\mathbf{Z}})\mathbf{W} = \phi(\widehat{\mathbf{Z}})\phi(\mathbf{X})^\top\mathbf{A} = \widehat{\mathbf{K}}\mathbf{A}$ where $\widehat{\mathbf{K}}\in\mathbb{R}^{n\times n}$ with its element $\widehat{k}_{ij} = \phi(\widehat{\mathbf{z}}_i)^\top\phi(\mathbf{x}_j)=\kappa(\widehat{\mathbf{z}}_i,\mathbf{x}_j)$.
With these notations in mind, we can obtain the following objective function:
\begin{align}
	\nonumber
	\min_{\mathbf{A},\mathbf{b}}\frac{1}{2}\left\|\mathbf{KA}+\mathbf{1}_n\mathbf{b}^\top-\mathbf{Y}\right\|_F^2  + \frac{\alpha}{2}\left\|\widetilde{\mathbf{K}}\mathbf{A}+\mathbf{1}_c\mathbf{b}^\top-\mathbf{T}\right\|_F^2&+\frac{\beta}{2}\text{tr}(\mathbf{A}^\top\mathbf{K}\mathbf{A})\\
	\label{eq16}
	&\quad\quad\quad + \frac{\gamma}{2}\text{tr}\Big(\mathbf{A}^\top(\mathbf{K}-\widehat{\mathbf{K}})^\top(\mathbf{K}-\widehat{\mathbf{K}})\mathbf{A}\Big),
\end{align}
where $\text{tr}(\cdot)$ denotes the trace operator, and we used its important property, i.e., $\left\|\mathbf{W}\right\|_F^2 = \text{tr}(\mathbf{W}^\top\mathbf{W})$. Since $\mathbf{W} = \phi(\mathbf{X})^\top\mathbf{A}$, $\left\|\mathbf{W}\right\|_F^2 =\text{tr}(\mathbf{A}^\top\phi(\mathbf{X})\phi(\mathbf{X})^\top\mathbf{A})=\text{tr}(\mathbf{A}^\top\mathbf{K}\mathbf{A})$. Similarly, the fourth term of problem~(\ref{eq16}) can also be derived in the same manner.
To solve problem~(\ref{eq16}), it is not hard to obtain the derivative with respect to $\mathbf{A}$ and $\mathbf{b}$:
\begin{align}
	\nabla_{\mathbf{A}} &= \mathbf{K}^\top(\mathbf{K}_\mathbf{A}+\mathbf{1}_n\mathbf{b}^\top-\mathbf{Y})+
	\alpha\widetilde{\mathbf{K}}^\top(\mathbf{K}\mathbf{A}+\mathbf{1}_c\mathbf{b}^\top-\mathbf{T})+\beta\mathbf{K}\mathbf{A}+\gamma(\mathbf{K}-\widehat{\mathbf{K}})^\top(\mathbf{K}-\widehat{\mathbf{K}})\mathbf{A},\\
	\nabla_{\mathbf{b}} &= (\mathbf{A}^\top\mathbf{K}^\top +\mathbf{b}\mathbf{1}^\top_n -\mathbf{Y}^\top)\mathbf{1}_n+\alpha(\mathbf{A}^\top\widetilde{\mathbf{K}}^\top+\mathbf{b}\mathbf{1}^\top_c-\mathbf{T}^\top)\mathbf{1}_c.
\end{align}
Setting $\nabla_{\mathbf{A}}$ and $\nabla_{\mathbf{b}}$ to 0, we can we can also obtain the closed-form solutions:
\begin{align}
	\nonumber
	\mathbf{A}&=\Big(\mathbf{K}^\top\mathbf{K}  +\alpha\widetilde{\mathbf{K}}^\top\widetilde{\mathbf{K}}+\beta\mathbf{K}+\gamma(\mathbf{K}-\widehat{\mathbf{K}})^\top(\mathbf{K}-\widehat{\mathbf{K}})
	-\frac{\mathbf{K}^\top\mathbf{1}_n+\alpha\widetilde{\mathbf{K}}^\top\mathbf{1}_c}{n+\alpha c}(\mathbf{1}_n^\top\mathbf{K} \\
	\label{eq20}
	&\quad\quad\quad\quad\quad\quad\quad\quad\quad\quad+\alpha\mathbf{1}^\top_{c}\widetilde{\mathbf{K}})\Big)^{-1}
	\Big(\mathbf{K}^\top\mathbf{Y}
	+\alpha\widetilde{\mathbf{K}}^\top\mathbf{T} - \frac{\mathbf{K}^\top\mathbf{1}_n + \alpha\widetilde{\mathbf{K}}^\top\mathbf{1}_c}{n+\alpha c} (\mathbf{1}_n^\top\mathbf{Y}+\alpha\mathbf{1}_c^\top\mathbf{T})\Big),\\
	\label{eq21}
	\mathbf{b} &= \frac{1}{n+\alpha c}\Big((\mathbf{Y}^\top - \mathbf{A}^\top\mathbf{K}^\top)\mathbf{1}_n 
	+ \alpha(\mathbf{T}^\top-\mathbf{A}^\top\widetilde{\mathbf{K}}^\top)\mathbf{1}_c\Big).
\end{align}

In this article, the Gaussian kernel function is adopted, i.e., $\kappa(\mathbf{x}_i,\mathbf{x}_j)=\exp(\frac{-\left\|\mathbf{x}_i-\mathbf{x}_j\right\|_2^2}{2\sigma^2})$,
where the kernel parameter $\sigma$ is empirically set to the averaged pairwise Euclidean distances of instances.
\begin{algorithm}[ht]
	\caption{The IMCC Algorithm}
	\label{alg1}
	\begin{algorithmic}[1]
		\REQUIRE\mbox{}\par
		$\mathcal{D}$: the multi-label training set $\mathcal{D}=\{(\mathbf{X}, \mathbf{Y})\}$\\
		$\alpha, \beta, \gamma$: the regularization hyperparameters\\
		$c$: the number of clusters\\
		$\mathbf{x}$: the unseen test instance
		\ENSURE \mbox{}\par
		$\mathbf{y}$: the predicted label vector for the test instance $\mathbf{x}$\\
		\item[]
		\STATE perform $k$-means clustering on $\mathbf{X}$;
		\STATE calculate cluster centers $\mathbf{Z}\in\mathbb{R}^{c\times d}$ according to Eq. (\ref{eq1});
		\STATE calculate the label vectors $\mathbf{T}\in[-1,+1]^{c\times q}$ of cluster centers according to Eq. (\ref{eq2});
		\STATE calculate the optimal solutions $\mathbf{A}^\star$ and $\mathbf{b}^\star$ according to Eq. (\ref{eq20}) and Eq. (\ref{eq21});
		\STATE return the predicted label vector $\mathbf{y}$ according to Eq. (\ref{eq22}).
	\end{algorithmic}
\end{algorithm}

\begin{table}[t]
	\centering
	\normalsize
	\caption{Characteristics of the benchmark multi-label datasets.}
	\setlength{\tabcolsep}{1.0mm}
	\label{dataset}
	\begin{tabular}{ccccccccc}
		\hline
		\hline
		Data set & $|D|$&$dim(D)$&$L(D)$&$F(D)$& $LCard(D)$ &$LDen(D)$  &$DL(D)$  &$PDL$  \\
		\hline
		cal500   & 502  & 68     & 174  & numeric & 26.044 &0.150  & 502 & 1.000  \\    
		image	 & 2000  & 294    & 5    & numeric & 1.236  & 0.247 &20  & 0.010  \\            
		scene  	& 2407 & 294    & 5    & numeric & 1.074 & 0.179 & 15 &0.006   \\             
		yeast    & 2417 & 103    & 14   & numeric & 4.237  &0.300  & 198 & 0.082 \\   
		enron    & 1702 &1001     	& 53    & nominal & 3.378 &0.064  &753  &0.442  \\   
		genbase    & 662 & 1185   & 27   & nominal & 1.252  & 0.046 & 32 &0.048  \\
		medical    & 978 & 1449   & 45   & nominal &1.245  & 0.028 & 94 &0.096  \\   
		\hline
		arts  	& 5000 & 462    & 26   & numeric & 1.636  & 0.063 &462  &0.924  \\    
		bibtex    & 7395 &1836     & 159   & nominal & 2.402  &0.015  & 2856 &0.386  \\    
		computer  & 5000 & 681    & 33  & nominal & 1.508 &0.046  &253  &0.051   \\    
		corel5k  & 5000 & 499    & 374  & nominal & 3.522  &0.009  &3175  &0.635  \\    
		education  & 5000 & 550    & 33  & nominal & 1.461 &0.443  &308  &0.062   \\    
		health  	& 5000 & 612    & 32  & nominal & 1.662 &0.052  &257  &0.051   \\   
		social  	& 5000 & 1047    & 39  & nominal & 1.283 &0.033  &226  &0.045   \\
		society   	& 5000 & 636    & 27  & nominal & 1.692  &0.063  &582  &0.116  \\
		\hline
		\hline
	\end{tabular}
\end{table}
\subsection{Test Phase}
Once the model parameters $\mathbf{A}$ and $\mathbf{b}$ are learned, we denote the optimal solutions as $\mathbf{A}^\star$ and $\mathbf{b}^\star$. Then, the predicted label vector $\mathbf{y}\in\{-1,+1\}^q$ of the test instance $\mathbf{x}$ is given as:
\begin{gather}
	\label{eq22}
	\mathbf{y} = \text{sign}(\sum_{i=1}^n\kappa(\mathbf{x},\mathbf{x}_i)\mathbf{a}_i^\star+\mathbf{b}^\star),
\end{gather}
where $\mathbf{a}_i^\star\in \mathbb{R}^{q}$ denote the $i$-th row of $A^{\star}$ and $\text{sign}(z)$ returns $+1$ if $z\geq 0$, otherwise $-1$. The pseudo code of IMCC 
is presented in Algorithm \ref{alg1}.

\section{Experiments}
In this section, we evaluate the performance of our proposed IMCC approach by comparing with multiple state-of-the-art approaches on a number of real-world multi-label datasets, using several widely used evaluation metrics.
\subsection{Experimental Setup}

\subsubsection{Datasets}
In order to get a persuasive comprehensive performance evaluation, we collect 15 real-world multi-label datasets for experimental analysis, all datasets  can  be downloaded from the MULAN\footnote{\url{http://mulan.sourceforge.net/datasets.html}} and MEKA\footnote{\url{https://sourceforge.net/projects/meka/files/Datasets/}}. For each data set $\mathcal{D}$, we denote by $|\mathcal{D}|$, $dim(\mathcal{D})$, $L(\mathcal{D})$, and $F(\mathcal{D})$ the number of examples, number of dimensions (features), number of class labels, and feature type, respectively. In addition, following~\cite{hou2016multi,zhang2018feature}, the properties of each data set are further characterized by several statistics, including label cardinality $LCard(\mathcal{D})$, label density $LDen(\mathcal{D})$, distinct label sets $DL(\mathcal{D})$ and proportion of distinct label set $PDL(\mathcal{D})$. The detailed definitions of these multi-label statistics can be found in~\cite{read2011classifier}. Table~\ref{dataset} reports the detailed information of all the datasets. According to $|\mathcal{D}|$, we divide the datasets into two parts: the regular-scale datasets for $|\mathcal{D}|<5000$ and the large-scale datasets for $|\mathcal{D}|\geq 5000$. For each data set, we randomly sample 80\% examples to form the training set, and the remaining 20\% examples belong to the test set. We repeat such sampling process for 10 times, and record the mean prediction value with the standard deviation.
\subsubsection{Comparing Algorithms} We compare our proposed approach IMCC with 6 state-of-the-art multi-label learning approaches. Each algorithm is configured with the suggested parameters according to the respective literature.
\begin{itemize}
	\item BRsvm~\cite{boutell2004learning}: It decomposes the multi-label classification problem into $L(\mathcal{D})$ independent binary (one-versus-rest) classification problems. The employed base model is binary SVM, which is trained by the \textsl{libsvm} toolbox~\cite{chang2011libsvm}.
	\item ECC~\cite{read2011classifier}: It is an ensemble of classifier chains, where the order of classifier chains is randomly generated. The employed base model is SVM, and the ensemble size is set to 10.
	\item MAHR~\cite{huang2012multi}: It uses a boosting approach and exploit label correlations by a hypothesis mechanism. The boosting round $T$ is set to $2\times dim(\mathcal{D})$.
	\item LIFT~\cite{zhang2014lift}: It constructs different features for different labels, train a binary SVM model for each label based on the label-specific features.
	\item LLSF~\cite{huang2016learning}: It learns label-specific features
	for multi-label learning. Parameters $\alpha$ and $\beta$ are searched in $\{2^{-10},2^{-9},\cdots,2^{10}\}$, and $\gamma$ is searched in $\{0.1,1,10\}$.
	\item JFSC~\cite{huang2018joint}: It performs joint feature selection and classification for multi-label learning. Parameters $\alpha$, $\beta$, and $\gamma$ are searched in $\{4^{-5},4^{-4},\cdots,4^5\}$, and $\eta$ is searched in $\{0.1, 1, 10\}$.
	\item IMCC: This is our proposed approach, which incorporates multiple cluster centers for multi-label learning. The regularization hyperparameters $\alpha$, $\beta$ and $\gamma$ are searched in $\{10^{-3},10^{-2},\cdots,10^{3}\}$, and the number of clusters $c$ is searched in $\{2^{3},\cdots,2^{7},2^{8}\}$.
\end{itemize}
For all the above approaches, the searched parameters are chosen by five-fold cross validation on the training set.

\subsubsection{Evaluation Metrics}

To comprehensively measure the performance of each multi-label learning approach, we adopt five widely used evaluation metrics, including \textsl{one error}, \textsl{hamming loss}, \textsl{ranking loss}, \textsl{coverage} and \textsl{average precision}. Note that for all the adopted multi-label evaluation metrics, their values are in the interval $[0,1]$. 
Given the train set $\mathcal{D}=\{\mathbf{x}_i,\mathbf{y}_i\}_{i=1}^n$ and the test set $\mathcal{D}_t=\{\mathbf{x}_i^t,\mathbf{y}_i^t\}_{i=1}^{n^t}$  where $\mathbf{x}_i, \mathbf{x}_i^t\in\mathbb{R}^d$ are the feature vector with $d$ dimensions (features) and $\mathbf{y}_i, \mathbf{y}_i^t\in\{-1,+1\}^q$ are the corresponding ground-truth label vector with the size of label space being $q$. The optimal model parameters $\mathbf{A}^\star=[\mathbf{a}_1^\star,\dots,\mathbf{a}_n^\star]^\top$ and $\mathbf{b}^\star$.
Then we can obtain $\mathbf{\hat{y}}_i^t = \text{sign}(\sum_{j}^{n}\kappa(\mathbf{x}_i^t,\mathbf{x}_j)\mathbf{a}_j^\star+\mathbf{b}^\star)$, the predicted label vector of $\mathbf{x}_i^t$.
\begin{itemize}
	\item \textbf{One error}: It evaluates the fraction that the label with the top-ranked predicted by the instance does not belong to its ground-truth relevant label set. The smaller the value of \textsl{one error}, the better performance of the classifier.
	\begin{align}
		\text{one-error} = \frac{1}{n^t}\sum_{i=1}^{n^t}\mathbb{I}[\mathbf{y}_{ij^{*}}^t=-1],
	\end{align}
	where $j^{*}=\arg\max_j\mathbf{\hat{y}}_i^t$, and $\mathbb{I}[z]$ returns 1 if $z$ holds and 0 otherwise.
	\item \textbf{Hamming loss}: It evaluates the fraction of instance label pairs which have been misclassified. The smaller the value of \textsl{hamming loss}, the better performance
	of the classifier.
	\begin{align}
		\text{hloss} = \frac{1}{n^t}\sum_{i=1}^{n^t}\frac{1}{q}\sum_{j=1}^q\mathbb{I}[y_{ij}^t\neq \hat{y}_{ij}^t],
	\end{align}
	\item \textbf{Rank loss}: It evaluates the average fraction of misordered label pairs. The smaller the value of \textsl{ranking loss}, the better performance of the classifier.
	\begin{align}
		\text{rloss} = \frac{1}{n^t}\sum_{i=1}^{n^t}\frac{1}{n_{i}^{1}n_{i}^{-1}}\sum_{j,k}^{q}\mathbb{I}[\tilde{y}_{ik}^t\leq \tilde{y}_{ij}^t],  
	\end{align}
	where $\mathbf{\tilde{y}}_i^t = \sum_{j}^{n}\kappa(\mathbf{x}_i^t,\mathbf{x}_j)\mathbf{a}_j^\star+\mathbf{b}^\star$, $1\leq j,k\leq q,\ y_{ij}^t=-1, y_{ik}^t=1$, $n_{i}^{1}$ denotes the number of positive label of $\mathbf{x}_i^t$, and $n_{i}^{-1}=q-n_{i}^{1}$ denotes the number of positive label. 
	\item \textbf{Coverage}: It evaluates how many steps are needed, on average, to move down the ranked label list of an instance so as to cover all its relevant labels. The smaller the value of \textsl{coverage}, the better performance of the classifier.
	\begin{align}
		\text{coverage} = \frac{1}{q}(\frac{1}{n^t}\sum_{i=1}^{n^t}\max_{j}rank(\tilde{y}_{ij}^t) - 1),
	\end{align}
	where $rank(\tilde{y}_{ij}^t)$  indicates the rank of $\tilde{y}_{ij}^t$ for $\mathbf{x}_i^t$.
	\item \textbf{Average precision}: It evaluates the average fraction of relevant labels ranked higher than a particular label. The larger the value of \textsl{average precision}, the better performance of the classifier.
	\begin{align}
		\text{average precision} = \frac{1}{n^t}\sum_{i=1}^{n^t}\frac{1}{n_{i}^1}\sum_{j}^q\frac{\left|\mathbf{R}(\mathbf{x}_i^t,\tilde{y}_{ij})\right|}{rank(\tilde{y}_{ij})},
	\end{align}
	where $\mathbf{R}(\mathbf{x}_i^t,\tilde{y}_{ij}^t)=\{\tilde{y}_{ik}^t|rank(\tilde{y}_{ik}^t)\leq rank(\tilde{y}_{ij}^t),\  y_{ik}^t=1,\  y_{ij}^t=1 \}$. 
\end{itemize}

\subsection{Experimental results}

\begin{table}[htbp]
	\centering
	\huge
	\caption{Predictive results of each algorithm (mean $\pm$ standard deviation) on the regular-scale datasets. The best results are highlighted, and the number in the brackets indicates the ranking of the algorithm.}
	\label{tab:regular}
	\resizebox{\textwidth}{90mm}{
		\begin{tabular}{ccccccccc}
			\hline
			Comparing &\multicolumn{7}{c}{One-error$\downarrow$}  \\
			\cline{2-8}
			algorithms &cal500	&image	&scene	&yeast	&enron	&genbase &medical \\
			\hline
			IMCC  &\textbf{0.116$\pm$0.024(1)}      &\textbf{0.253$\pm$0.021(1)}      &\textbf{0.179$\pm$0.017(1)}      &\textbf{0.210$\pm$0.015(1)}      &0.230$\pm$0.014(2)      &\textbf{0.002$\pm$0.005(1)}      &\textbf{0.117$\pm$0.018(1)}    \\
			BRsvm  &0.119$\pm$0.025(3)      &0.312$\pm$0.018(3)      &0.260$\pm$0.022(6)      &0.225$\pm$0.016(4)      &0.285$\pm$0.023(6)      &0.101$\pm$0.313(5)      &0.235$\pm$0.044(7)    \\
			ECC  &0.118$\pm$0.023(2)      &0.321$\pm$0.020(4)      &0.241$\pm$0.016(3)      &0.236$\pm$0.020(5)      &0.298$\pm$0.019(7)      &0.101$\pm$0.314(5)      &0.223$\pm$0.067(5)    \\
			MAHR  &0.186$\pm$0.092(7)      &0.306$\pm$0.016(2)      &0.231$\pm$0.010(2)      &0.238$\pm$0.017(6)      &0.265$\pm$0.016(5)      &0.002$\pm$0.003(2)      &0.146$\pm$0.027(4)    \\
			LLSF  &0.122$\pm$0.023(5)      &0.331$\pm$0.021(7)      &0.254$\pm$0.015(5)      &0.358$\pm$0.023(7)      &\textbf{0.226$\pm$0.017(1)}      &0.002$\pm$0.003(3)      &0.126$\pm$0.016(2)    \\
			JFSC  &0.119$\pm$0.023(4)      &0.329$\pm$0.026(6)      &0.270$\pm$0.011(7)      &0.217$\pm$0.011(2)      &0.239$\pm$0.014(3)      &0.004$\pm$0.005(4)      &0.143$\pm$0.022(3)    \\
			LIFT  &0.122$\pm$0.024(5)      &0.326$\pm$0.024(5)      &0.241$\pm$0.019(3)      &0.221$\pm$0.013(3)      &0.251$\pm$0.022(4)      &0.101$\pm$0.314(5)      &0.230$\pm$0.051(6)    \\
			
			\hline
			Comparing&\multicolumn{7}{c}{Hamming loss$\downarrow$}  \\
			\cline{2-8}
			algorithms&cal500	&image	&scene	&yeast	&enron	&genbase &medical  \\
			\hline
			IMCC  &\textbf{0.137$\pm$0.003(1)}      &\textbf{0.148$\pm$0.009(1)}      &\textbf{0.077$\pm$0.004(1)}      &\textbf{0.191$\pm$0.005(1)}      &\textbf{0.046$\pm$0.002(1)}      &0.002$\pm$0.001(4)      &\textbf{0.010$\pm$0.001(1)}    \\
			BRsvm  &\textbf{0.137$\pm$0.003(1)}      &0.181$\pm$0.011(3)      &0.105$\pm$0.004(5)      &0.199$\pm$0.005(2)      &0.051$\pm$0.002(4)      &0.005$\pm$0.012(5)      &0.013$\pm$0.007(5)    \\
			ECC  &0.154$\pm$0.004(7)      &0.256$\pm$0.011(7)      &0.155$\pm$0.009(7)      &0.249$\pm$0.005(6)      &0.061$\pm$0.002(7)      &0.005$\pm$0.012(5)      &0.015$\pm$0.031(7)    \\
			MAHR  &0.141$\pm$0.003(6)      &0.171$\pm$0.007(2)      &0.091$\pm$0.003(2)      &0.207$\pm$0.005(5)      &0.051$\pm$0.001(4)      &\textbf{0.001$\pm$0.001(1)}      &\textbf{0.010$\pm$0.001(1)}    \\
			LLSF  &0.138$\pm$0.003(3)      &0.181$\pm$0.009(3)      &0.103$\pm$0.003(4)      &0.301$\pm$0.004(7)      &\textbf{0.046$\pm$0.002(1)}      &\textbf{0.001$\pm$0.001(1)}      &\textbf{0.010$\pm$0.001(1)}    \\
			JFSC  &0.138$\pm$0.003(4)      &0.186$\pm$0.008(6)      &0.118$\pm$0.004(6)      &0.199$\pm$0.005(2)      &0.052$\pm$0.002(6)      &\textbf{0.001$\pm$0.001(1)}      &\textbf{0.010$\pm$0.001(1)}    \\
			LIFT  &0.139$\pm$0.003(5)      &\textbf{0.181$\pm$0.010(1)}      &0.098$\pm$0.004(3)      &0.199$\pm$0.005(2)      &0.047$\pm$0.001(3)      &0.005$\pm$0.012(5)      &0.013$\pm$0.007(5)    \\

			\hline
			Comparing&\multicolumn{7}{c}{Ranking loss$\downarrow$}  \\
			\cline{2-8}
			algorithms&cal500	&image	&scene	&yeast	&enron	&genbase &medical  \\
			
			\hline
			IMCC  &\textbf{0.181$\pm$0.005(1)}      &\textbf{0.137$\pm$0.010(1)}      &\textbf{0.061$\pm$0.007(1)}      &\textbf{0.157$\pm$0.005(1)}      &\textbf{0.074$\pm$0.006(1)}      &\textbf{0.001$\pm$0.003(1)}      &0.018$\pm$0.005(2)    \\
			BRsvm  &0.183$\pm$0.004(2)      &0.169$\pm$0.011(4)      &0.089$\pm$0.007(5)      &0.169$\pm$0.003(3)      &0.084$\pm$0.008(4)      &0.009$\pm$0.013(6)      &0.026$\pm$0.010(6)    \\
			ECC  &0.189$\pm$0.004(6)      &0.165$\pm$0.009(2)      &0.081$\pm$0.005(3)      &0.171$\pm$0.006(4)      &0.084$\pm$0.007(4)      &0.009$\pm$0.013(6)      &0.025$\pm$0.010(5)    \\
			MAHR  &0.275$\pm$0.010(7)      &0.165$\pm$0.008(2)      &0.083$\pm$0.005(4)      &0.181$\pm$0.005(6)      &0.129$\pm$0.006(7)      &0.005$\pm$0.003(4)      &0.027$\pm$0.008(7)    \\
			LLSF  &0.187$\pm$0.007(5)      &0.178$\pm$0.014(7)      &0.091$\pm$0.005(6)      &0.341$\pm$0.007(7)      &0.081$\pm$0.008(2)      &0.002$\pm$0.002(2)      &\textbf{0.017$\pm$0.005(1)}    \\
			JFSC  &0.184$\pm$0.006(4)      &0.175$\pm$0.015(6)      &0.096$\pm$0.005(7)      &0.171$\pm$0.005(4)      &0.098$\pm$0.007(6)      &\textbf{0.001$\pm$0.001(1)}      &0.019$\pm$0.006(3)    \\
			LIFT  &0.183$\pm$0.004(2)      &0.171$\pm$0.013(5)      &0.078$\pm$0.004(2)      &0.168$\pm$0.005(2)      &0.081$\pm$0.007(2)      &0.008$\pm$0.014(5)      &0.024$\pm$0.010(4)    \\

			\hline
			Comparing&\multicolumn{7}{c}{Coverage$\downarrow$}  \\
			\cline{2-8}
			algorithms&cal500	&image	&scene	&yeast	&enron	&genbase &medical \\
			\hline
			
			IMCC  &0.747$\pm$0.014(2)      &\textbf{0.167$\pm$0.013(1)}      &\textbf{0.066$\pm$0.007(1)}      &\textbf{0.441$\pm$0.006(1)}      &\textbf{0.221$\pm$0.017(1)}      &\textbf{0.011$\pm$0.006(1)}      &\textbf{0.028$\pm$0.008(1)}    \\
			BRsvm  &0.751$\pm$0.014(4)      &0.191$\pm$0.012(4)      &0.089$\pm$0.006(5)      &0.458$\pm$0.006(4)      &0.235$\pm$0.021(5)      &0.022$\pm$0.014(5)      &0.041$\pm$0.013(6)    \\
			ECC  &0.765$\pm$0.013(6)      &0.187$\pm$0.010(2)      &0.081$\pm$0.004(3)      &0.455$\pm$0.008(2)      &0.228$\pm$0.018(3)      &0.022$\pm$0.014(5)      &0.039$\pm$0.012(5)    \\
			MAHR  &0.894$\pm$0.012(7)      &0.189$\pm$0.008(3)      &0.084$\pm$0.004(4)      &0.477$\pm$0.007(6)      &0.339$\pm$0.020(7)      &0.013$\pm$0.002(3)      &0.041$\pm$0.010(6)    \\
			LLSF  &0.747$\pm$0.015(2)      &0.194$\pm$0.015(5)      &0.092$\pm$0.004(6)      &0.627$\pm$0.009(7)      &0.222$\pm$0.019(2)      &0.013$\pm$0.003(3)      &\textbf{0.028$\pm$0.008(1)}    \\
			JFSC  &\textbf{0.742$\pm$0.014(1)}      &0.194$\pm$0.015(5)      &0.092$\pm$0.005(6)      &0.455$\pm$0.007(2)      &0.265$\pm$0.017(6)      &\textbf{0.011$\pm$0.002(1)}      &0.029$\pm$0.009(3)    \\
			LIFT  &0.751$\pm$0.017(4)      &0.194$\pm$0.015(5)      &0.079$\pm$0.003(2)      &0.461$\pm$0.007(5)      &0.228$\pm$0.018(3)      &0.022$\pm$0.014(5)      &0.038$\pm$0.011(4)    \\

			\hline
			Comparing&\multicolumn{7}{c}{Average precision$\uparrow$}  \\
			\cline{2-8}
			algorithms&cal500	&image	&scene	&yeast	&enron	&genbase &medical  \\
			\hline
			
			IMCC  &\textbf{0.505$\pm$0.005(1)}      &\textbf{0.834$\pm$0.012(1)}      &\textbf{0.893$\pm$0.010(1)}      &\textbf{0.777$\pm$0.008(1)}      &\textbf{0.704$\pm$0.013(1)}      &\textbf{0.997$\pm$0.004(1)}      &\textbf{0.912$\pm$0.012(1)}    \\
			BRsvm  &0.501$\pm$0.006(2)      &0.797$\pm$0.011(3)      &0.847$\pm$0.012(5)      &0.762$\pm$0.008(3)      &0.657$\pm$0.016(4)      &0.944$\pm$0.152(6)      &0.841$\pm$0.132(7)    \\
			ECC  &0.491$\pm$0.003(6)      &0.797$\pm$0.011(3)      &0.857$\pm$0.008(4)      &0.756$\pm$0.011(5)      &0.657$\pm$0.013(4)      &0.944$\pm$0.152(6)      &0.852$\pm$0.134(5)    \\
			MAHR  &0.441$\pm$0.010(7)      &0.801$\pm$0.008(2)      &0.861$\pm$0.006(2)      &0.745$\pm$0.009(6)      &0.641$\pm$0.013(7)      &0.994$\pm$0.003(4)      &0.892$\pm$0.018(4)    \\
			LLSF  &0.501$\pm$0.010(2)      &0.789$\pm$0.014(5)      &0.847$\pm$0.007(5)      &0.617$\pm$0.007(7)      &0.703$\pm$0.015(2)      &0.996$\pm$0.003(2)      &0.908$\pm$0.009(2)    \\
			JFSC  &0.501$\pm$0.007(2)      &0.789$\pm$0.016(5)      &0.836$\pm$0.007(7)      &0.762$\pm$0.008(3)      &0.643$\pm$0.013(6)      &0.996$\pm$0.003(2)      &0.899$\pm$0.013(3)    \\
			LIFT  &0.496$\pm$0.006(5)      &0.789$\pm$0.015(5)      &0.859$\pm$0.010(3)      &0.766$\pm$0.007(2)      &0.684$\pm$0.013(3)      &0.947$\pm$0.153(5)      &0.848$\pm$0.023(6)    \\
			\hline
			\hline
			
		\end{tabular}
	}
\end{table}

\begin{table}[htbp]
	\centering
	\huge
	\caption{Predictive results of each algorithm (mean $\pm$ standard deviation) on the large-scale datasets. The best results are highlighted, and the number in the brackets indicates the ranking of the algorithm.}
	\label{tab:large1}
	\resizebox{\textwidth}{90mm}{
		\begin{tabular}{ccccccccc}
			\hline
			\hline
			Comparing &\multicolumn{7}{c}{One-error$\downarrow$}  \\
			\cline{2-9}
			algorithms &arts  &bibtex		&computer		&corel5k		&education		&health		&social		&society	
			\\
			\hline
			IMCC  &\textbf{0.456$\pm$0.013(1)}      &0.361$\pm$0.008(4)      &\textbf{0.333$\pm$0.014(1)}      &0.661$\pm$0.009(2)      &0.462$\pm$0.016(2)      &0.254$\pm$0.011(2)      &\textbf{0.272$\pm$0.004(1)}      &\textbf{0.386$\pm$0.018(1)}    \\
			BRsvm  &\textbf{0.456$\pm$0.014(1)}      &0.403$\pm$0.015(7)      &0.407$\pm$0.209(4)      &0.702$\pm$0.105(5)      &\textbf{0.271$\pm$0.031(1)}      &0.468$\pm$0.367(5)      &0.409$\pm$0.311(5)      &0.446$\pm$0.195(4)    \\
			ECC  &0.482$\pm$0.010(5)      &0.394$\pm$0.012(6)      &0.413$\pm$0.206(6)      &0.718$\pm$0.099(6)      &0.571$\pm$0.226(5)      &0.473$\pm$0.364(6)      &0.414$\pm$0.309(6)      &0.452$\pm$0.193(6)    \\
			MAHR  &0.548$\pm$0.011(7)      &0.371$\pm$0.005(5)      &0.409$\pm$0.014(5)      &0.907$\pm$0.008(7)      &0.603$\pm$0.021(7)      &0.321$\pm$0.015(4)      &0.328$\pm$0.007(4)      &0.446$\pm$0.015(4)    \\
			LLSF  &0.461$\pm$0.011(4)      &\textbf{0.349$\pm$0.004(1)}      &0.337$\pm$0.017(2)      &\textbf{0.624$\pm$0.011(1)}      &0.466$\pm$0.013(3)      &\textbf{0.246$\pm$0.015(1)}      &0.273$\pm$0.008(2)      &0.394$\pm$0.017(2)    \\
			JFSC  &0.512$\pm$0.012(6)      &0.358$\pm$0.007(3)      &0.381$\pm$0.014(3)      &0.675$\pm$0.008(3)      &0.515$\pm$0.022(4)      &0.296$\pm$0.009(3)      &0.323$\pm$0.008(3)      &0.423$\pm$0.018(3)    \\
			LIFT  &\textbf{0.456$\pm$0.011(1)}      &0.355$\pm$0.011(2)      &0.413$\pm$0.206(6)      &0.683$\pm$0.112(4)      &0.581$\pm$0.221(6)      &0.478$\pm$0.361(7)      &0.427$\pm$0.302(7)      &0.469$\pm$0.187(7)    \\
			\hline
			Comparing &\multicolumn{7}{c}{Hamming loss$\downarrow$}  \\
			\cline{2-9}
			algorithms &arts  &bibtex &computer	&corel5k &education &health &social &society	\\
			\hline
			IMCC  &0.057$\pm$0.001(3)      &0.013$\pm$0.0(3)      &\textbf{0.033$\pm$0.002(1)}      &\textbf{0.009$\pm$0.001(1)}      &\textbf{0.038$\pm$0.001(1)}      &\textbf{0.033$\pm$0.001(1)}      &\textbf{0.021$\pm$0.001(1)}      &\textbf{0.051$\pm$0.001(1)}    \\
			BRsvm  &\textbf{0.054$\pm$0.001(1)}      &0.013$\pm$0.0(3)      &0.036$\pm$0.009(3)      &0.011$\pm$0.001(5)      &0.199$\pm$0.009(7)      &0.041$\pm$0.015(5)      &0.024$\pm$0.011(5)      &0.055$\pm$0.012(4)    \\
			ECC  &0.077$\pm$0.004(7)      &0.014$\pm$0.0(6)      &0.046$\pm$0.009(7)      &0.011$\pm$0.001(5)      &0.059$\pm$0.013(6)      &0.048$\pm$0.015(6)      &0.031$\pm$0.011(7)      &0.061$\pm$0.012(6)    \\
			MAHR  &0.057$\pm$0.001(3)      &0.013$\pm$0.0(3)      &0.037$\pm$0.002(5)      &\textbf{0.009$\pm$0.001(1)}      &0.041$\pm$0.001(4)      &0.038$\pm$0.002(4)      &0.022$\pm$0.001(3)      &0.056$\pm$0.001(5)    \\
			LLSF  &0.057$\pm$0.001(3)      &\textbf{0.012$\pm$0.0(1)}      &0.034$\pm$0.001(2)      &\textbf{0.009$\pm$0.001(1)}      &\textbf{0.038$\pm$0.001(1)}      &\textbf{0.033$\pm$0.001(1)}      &\textbf{0.021$\pm$0.001(1)}      &0.052$\pm$0.001(2)    \\
			JFSC  &0.057$\pm$0.001(3)      &0.017$\pm$0.0(7)      &0.036$\pm$0.002(3)      &\textbf{0.009$\pm$0.001(1)}      &0.039$\pm$0.001(3)      &0.036$\pm$0.001(3)      &0.022$\pm$0.001(3)      &0.053$\pm$0.001(3)    \\
			LIFT  &\textbf{0.054$\pm$0.001(1)}      &\textbf{0.012$\pm$0.0(1)}      &0.037$\pm$0.009(5)      &0.011$\pm$0.001(5)      &0.044$\pm$0.013(5)      &0.048$\pm$0.016(6)      &0.024$\pm$0.011(5)      &0.061$\pm$0.012(6)    \\
			\hline
			Comparing &\multicolumn{7}{c}{Ranking loss$\downarrow$}  \\
			\cline{2-9}
			algorithms &arts  &bibtex &computer	&corel5k &education &health &social &society	\\
			\hline
			IMCC  &\textbf{0.111$\pm$0.003(1)}      &\textbf{0.063$\pm$0.002(1)}      &0.077$\pm$0.004(4)      &\textbf{0.111$\pm$0.002(1)}      &\textbf{0.072$\pm$0.004(1)}      &\textbf{0.046$\pm$0.003(1)}      &0.052$\pm$0.004(2)      &0.126$\pm$0.005(3)    \\
			BRsvm  &0.114$\pm$0.004(2)      &0.085$\pm$0.001(6)      &0.071$\pm$0.010(2)      &0.123$\pm$0.003(3)      &0.156$\pm$0.012(6)      &0.049$\pm$0.016(3)      &0.052$\pm$0.012(2)      &0.123$\pm$0.012(2)    \\
			ECC  &0.115$\pm$0.004(4)      &0.083$\pm$0.002(4)      &\textbf{0.068$\pm$0.010(1)}      &0.122$\pm$0.003(2)      &0.076$\pm$0.015(2)      &0.048$\pm$0.016(2)      &\textbf{0.049$\pm$0.011(1)}      &\textbf{0.121$\pm$0.012(1)}    \\
			MAHR  &0.201$\pm$0.010(7)      &0.094$\pm$0.004(7)      &0.125$\pm$0.006(7)      &0.266$\pm$0.018(7)      &0.209$\pm$0.012(7)      &0.077$\pm$0.006(7)      &0.095$\pm$0.006(7)      &0.211$\pm$0.008(7)    \\
			LLSF  &0.121$\pm$0.004(5)      &0.069$\pm$0.002(2)      &0.089$\pm$0.005(5)      &0.126$\pm$0.004(5)      &0.081$\pm$0.004(4)      &0.062$\pm$0.003(5)      &0.061$\pm$0.005(5)      &0.137$\pm$0.005(5)    \\
			JFSC  &0.122$\pm$0.004(6)      &0.083$\pm$0.003(4)      &0.095$\pm$0.004(6)      &0.138$\pm$0.002(6)      &0.081$\pm$0.005(4)      &0.069$\pm$0.005(6)      &0.078$\pm$0.006(6)      &0.146$\pm$0.006(6)    \\
			LIFT  &0.114$\pm$0.004(3)      &0.074$\pm$0.002(3)      &0.074$\pm$0.011(3)      &0.123$\pm$0.003(3)      &0.078$\pm$0.015(3)      &0.051$\pm$0.016(4)      &0.052$\pm$0.011(2)      &0.126$\pm$0.013(3)    \\
			\hline
			Comparing &\multicolumn{7}{c}{Coverage$\downarrow$}  \\
			\cline{2-9}
			algorithms &arts  &bibtex &computer	&corel5k &education &health &social &society	\\
			\hline
			IMCC  &\textbf{0.173$\pm$0.004(1)}      &\textbf{0.124$\pm$0.003(1)}      &0.118$\pm$0.006(4)      &\textbf{0.269$\pm$0.006(1)}      &0.105$\pm$0.005(2)      &0.096$\pm$0.006(4)      &0.081$\pm$0.006(4)      &0.207$\pm$0.007(4)    \\
			BRsvm  &0.174$\pm$0.006(3)      &0.158$\pm$0.003(6)      &0.107$\pm$0.010(2)      &0.289$\pm$0.006(4)      &0.291$\pm$0.015(7)      &0.089$\pm$0.015(2)      &0.071$\pm$0.011(2)      &0.189$\pm$0.014(2)    \\
			ECC  &\textbf{0.173$\pm$0.007(1)}      &0.156$\pm$0.003(5)      &\textbf{0.105$\pm$0.009(1)}      &0.287$\pm$0.006(3)      &\textbf{0.103$\pm$0.015(1)}      &\textbf{0.088$\pm$0.014(1)}      &\textbf{0.068$\pm$0.011(1)}      &\textbf{0.188$\pm$0.014(1)}    \\
			MAHR  &0.279$\pm$0.012(7)      &0.171$\pm$0.004(7)      &0.174$\pm$0.007(7)      &0.515$\pm$0.027(7)      &0.264$\pm$0.014(6)      &0.136$\pm$0.009(7)      &0.128$\pm$0.007(7)      &0.307$\pm$0.009(7)    \\
			LLSF  &0.189$\pm$0.006(6)      &0.132$\pm$0.004(2)      &0.131$\pm$0.006(5)      &0.281$\pm$0.006(2)      &0.119$\pm$0.005(5)      &0.121$\pm$0.005(5)      &0.091$\pm$0.006(5)      &0.216$\pm$0.008(5)    \\
			JFSC  &0.184$\pm$0.006(5)      &0.151$\pm$0.004(4)      &0.142$\pm$0.005(6)      &0.319$\pm$0.004(6)      &0.114$\pm$0.007(4)      &0.133$\pm$0.008(6)      &0.109$\pm$0.007(6)      &0.231$\pm$0.011(6)    \\
			LIFT  &0.174$\pm$0.006(3)      &0.141$\pm$0.003(3)      &0.111$\pm$0.010(3)      &0.289$\pm$0.006(4)      &0.106$\pm$0.015(3)      &0.091$\pm$0.015(3)      &0.071$\pm$0.011(2)      &0.191$\pm$0.014(3)    \\
			\hline
			Comparing &\multicolumn{7}{c}{Average precision$\uparrow$}  \\
			\cline{2-9}
			algorithms &arts  &bibtex &computer	&corel5k &education &health &social &society	\\
			\hline
			IMCC  &\textbf{0.634$\pm$0.008(1)}      &0.608$\pm$0.006(2)      &\textbf{0.723$\pm$0.010(1)}      &0.296$\pm$0.002(2)      &0.648$\pm$0.013(2)      &\textbf{0.795$\pm$0.008(1)}      &\textbf{0.786$\pm$0.007(1)}      &\textbf{0.648$\pm$0.010(1)}    \\
			BRsvm  &0.627$\pm$0.009(2)      &0.538$\pm$0.010(7)      &0.685$\pm$0.099(3)      &0.271$\pm$0.027(4)      &\textbf{0.807$\pm$0.014(1)}      &0.695$\pm$0.167(6)      &0.719$\pm$0.155(4)      &0.622$\pm$0.086(3)    \\
			ECC  &0.617$\pm$0.007(5)      &0.548$\pm$0.008(6)      &0.685$\pm$0.099(3)      &0.265$\pm$0.027(5)      &0.591$\pm$0.115(5)      &0.698$\pm$0.168(5)      &0.719$\pm$0.153(4)      &0.619$\pm$0.087(5)    \\
			MAHR  &0.524$\pm$0.008(7)      &0.574$\pm$0.005(5)      &0.635$\pm$0.010(7)      &0.099$\pm$0.005(7)      &0.481$\pm$0.016(7)      &0.725$\pm$0.009(4)      &0.715$\pm$0.007(6)      &0.561$\pm$0.010(7)    \\
			LLSF  &0.627$\pm$0.007(2)      &\textbf{0.613$\pm$0.005(1)}      &0.714$\pm$0.011(2)      &\textbf{0.305$\pm$0.008(1)}      &0.642$\pm$0.010(3)      &0.786$\pm$0.008(2)      &0.780$\pm$0.008(2)      &0.639$\pm$0.010(2)    \\
			JFSC  &0.597$\pm$0.007(6)      &0.593$\pm$0.006(3)      &0.685$\pm$0.009(3)      &0.261$\pm$0.003(6)      &0.615$\pm$0.014(4)      &0.761$\pm$0.006(3)      &0.751$\pm$0.007(3)      &0.622$\pm$0.010(3)    \\
			LIFT  &0.627$\pm$0.007(2)      &0.585$\pm$0.007(4)      &0.678$\pm$0.098(6)      &0.281$\pm$0.028(3)      &0.582$\pm$0.113(6)      &0.688$\pm$0.164(7)      &0.708$\pm$0.152(7)      &0.609$\pm$0.085(6)    \\
			
			\hline
			\hline
			
		\end{tabular}
	}
\end{table}

Tables \ref{tab:regular} and \ref{tab:large1} report the detailed experimental results of each algorithm on regular-scale and large-scale datasets, respectively. For the two tables, the best results are highlighted (in boldface), and the number in each bracket indicates the ranking of this algorithm.

\begin{table}[htbp]
	\centering
	\normalsize
	\setlength{\abovecaptionskip}{0.1cm}   
	\setlength{\belowcaptionskip}{0.2cm}
	\caption{Friedman statistics $F_F$ according to each evaluation metric and the critical value at 0.05 significance level (comparing algorithms $k=7$ and datasets $N=15$).}
	\label{tab:FF}
	\begin{tabular}{c|c|c} 
		\hline
		\hline
		Evaluation metric & $F_F$  &critical value ($\alpha$ = 0.05)  \\
		\hline
		One-error         &4.57 &\multirow{5}{*}{2.209} \\
		Hamming loss       &6.06 \\
		Ranking loss    &13.74 \\
		Coverage      &6.76\\
		Average precision   &11.45\\
		\hline
		\hline
	\end{tabular}
\end{table}

In order to further systematically analyze the relative performance of each comparing algorithm, we use the popular statistical test - \textsl{Friedman test}~\cite{demvsar2006statistical} for the comparison studies of multiple algorithms on a number of datasets, with respect to each evaluation metric. Specifically, given $k$ algorithms to be compared on $N$ datasets, 
and the $i$-th algorithm's average ranking on all the datasets is denoted by $r_i$. Note that mean ranks are shared in case of the performance of the algorithms are equal. Based on the null hypothesis that the performance of all algorithms is equal, the Friedman statistics $F_F$ is calculated by: $F_{F}= (N-1)\chi^{2}_{F}/(N(k-1)-\chi^{2}_{F}),$
where the $\chi^{2}_{F}$ is distributed to the $\chi^{2}$ distribution with $(k-1)$ degrees of freedom:

\begin{gather}
	\chi^{2}_{F} = \frac{12N}{k(k+1)}\Big[\sum^{k}_{i=1}r^{2}_{i}-\frac{k(k+1)^{2}}{4}\Big].
\end{gather}

In this article, the number of comparing algorithms $k=7$, the number of datasets $N=15$. Table \ref{tab:FF} summarizes the Friedman statistics $F_F$ according to each evaluation metric and the critical value at 0.05 significance level. As shown in Table \ref{tab:FF}, the equal hypothesis is obviously rejected at the significance level $\alpha = 0.05$. Consequently, the \textsl{post-hoc test} \cite{demvsar2006statistical} is used for further analysis. It makes sense to employ \textsl{Nemenyi test} \cite{demvsar2006statistical} to indicate whether our proposed IMCC approach achieves a superior performance to the comparing algorithms by treating IMCC as the control algorithm. The significant differences between IMCC and other algorithms can be determined by comparing their average ranking with the \textit{Critical Difference} (CD) \cite{demvsar2006statistical} ($\text{CD} = q_{\alpha}\sqrt{k(k+1)/6N}$).

\begin{figure}[htbp]
	\centering
	\setlength{\abovecaptionskip}{0.1cm}   
	\setlength{\belowcaptionskip}{-0.1cm}  
	\subfigure[\textsl{One-error}]{\includegraphics[width=0.45\textwidth]{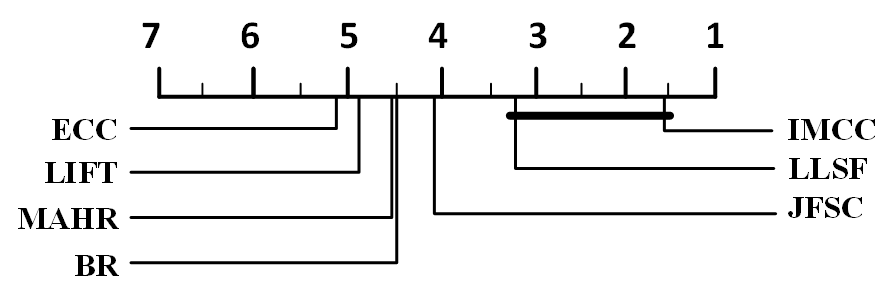}}
	\\
	\subfigure[\textsl{Hamming loss}]{\includegraphics[width=0.45\textwidth]{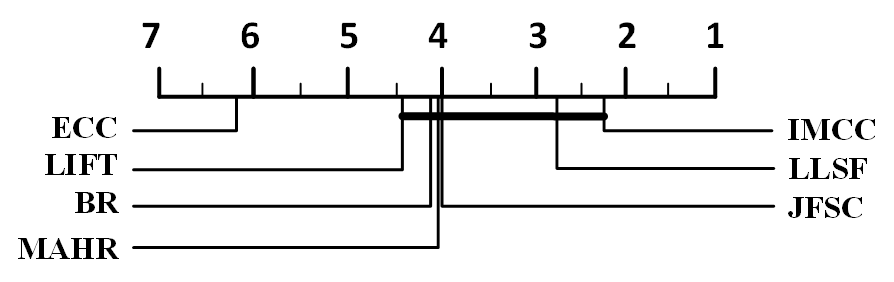}}
	\subfigure[\textit{Ranking loss}]{\includegraphics[width=0.45\textwidth]{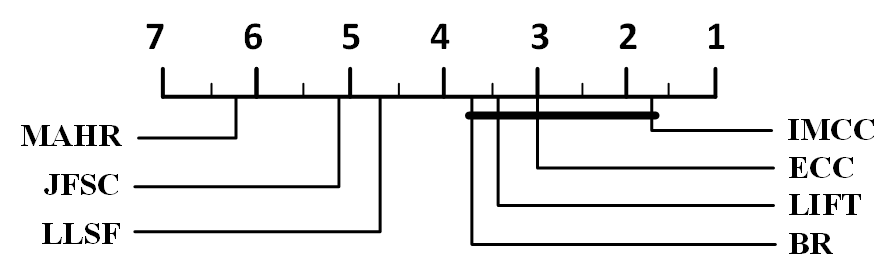}}
	\\
	\subfigure[\textsl{Coverage}]{\includegraphics[width=0.45\textwidth]{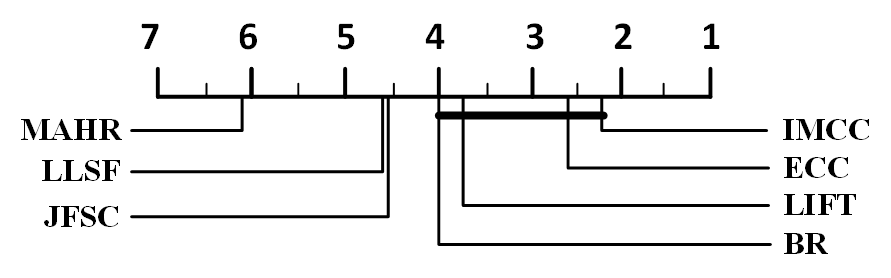}}
	\subfigure[\textsl{Average precision}]{\includegraphics[width=0.45\textwidth]{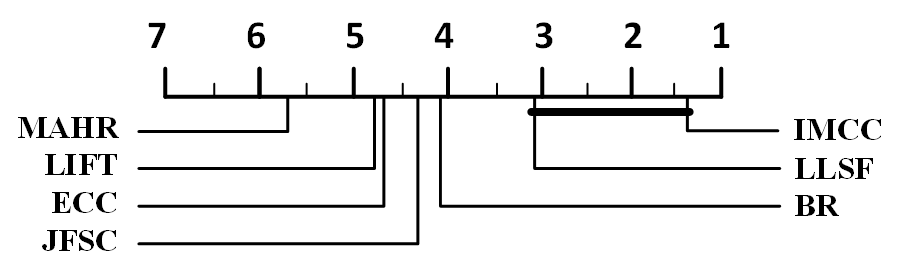}}
	\caption{Comparison of IMCC (control algorithm) against other comparing algorithms based on the \textsl{Nemenyi test}. Algorithms not connected with IMCC in the CD diagram are considered to have significantly different performance from IMCC.}
	\label{fig:CD}
\end{figure}

Given $\alpha = 0.05$, $k = 7$ and $N = 15$, for the \textsl{Nemenyi test}, $q_{\alpha} = 2.949$, we can obtain $\text{CD} = 2.3261$.
The performance of an algorithm is considered to be significantly different from that of IMCC if their average ranking over all datasets differs at least one CD. Figure \ref{fig:CD} shows the CD diagrams on each evaluation metric. In Figure\ref{fig:CD}, the comparison algorithm is connected to the IMCC if their average rank is within one CD to that of IMCC. Otherwise, there exists significantly different performance between IMCC and a comparing algorithm if the algorithm is not connected with the IMCC. 

Based on the above experimental results, the following observations can be made:
\begin{itemize}
	\item As shown in Table \ref{tab:regular} and Table \ref{tab:large1}, IMCC ranked first on all evaluation metrics on the four datasets including \textsl{image}, \textsl{scene}, \textsl{yeast} and \textsl{arts}). This is because these four datasets are regular-scale datasets, which have limited number of examples, and IMCC can achieve great performance on regular-scale datasets due to data augmentation.
	\item From both Table \ref{tab:regular} and Table \ref{tab:large1}, we can observe that across all evaluation metrics and on all the fifteen datasets, IMCC ranks first on all the fifteen datasets in 72.00\% cases, and ranks top three in 89.33\% cases. It is also worthy noting that IMCC ranks first in 88.57\% cases on the regular-scale datasets (Table \ref{tab:regular}) while IMCC ranks first in 55.50\% cases on the large-scale datasets (Table \ref{tab:large1}). These results indicate that IMCC is superior to other comparing algorithms in most cases and IMCC tends to work better on regular-scale datasets. Such observation accords with the widely-accepted intuition that the data augmentation approach is normally more helpful to the regular-scale datasets compared with the large-scale datasets. As the large-scale datasets may provide relatively adequate training examples, data augmentation might be not much useful in this case. Despite this, IMCC still achieves competitive performance against other state-of-the-art approaches on the large-scale datasets.
	
	\item From Figure~\ref{fig:CD}, we can observe that, in all cases, IMCC achieves the best performance compared to all algorithms. It is also worthy noting that IMCC significantly outperforms each comparing algorithm on at least two evaluation metrics. Moreover, on the \textsl{one-error} and \textsl{average precision} metrics, only LLSF is competitive against IMCC (i.e., IMCC significantly outperforms the other five algorithms on the two evaluation metrics). 
\end{itemize}
\begin{figure}[tbp]
	\centering
	\subfigure[Varying $\alpha$ and $\gamma$ on \textsl{enron}]{
		\includegraphics[width=0.3\textwidth]{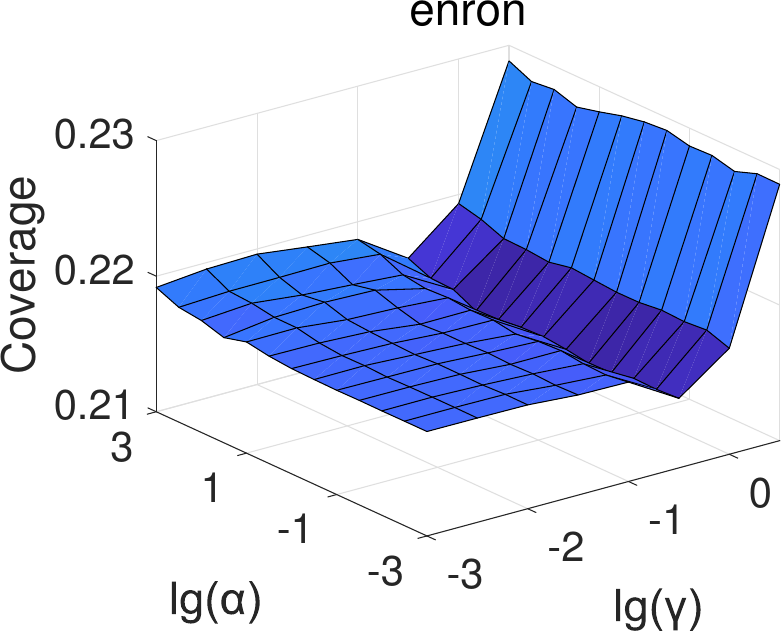}}
	\subfigure[Varying $\alpha$ and $\gamma$ on \textsl{yeast}]{
		\includegraphics[width=0.3\textwidth]{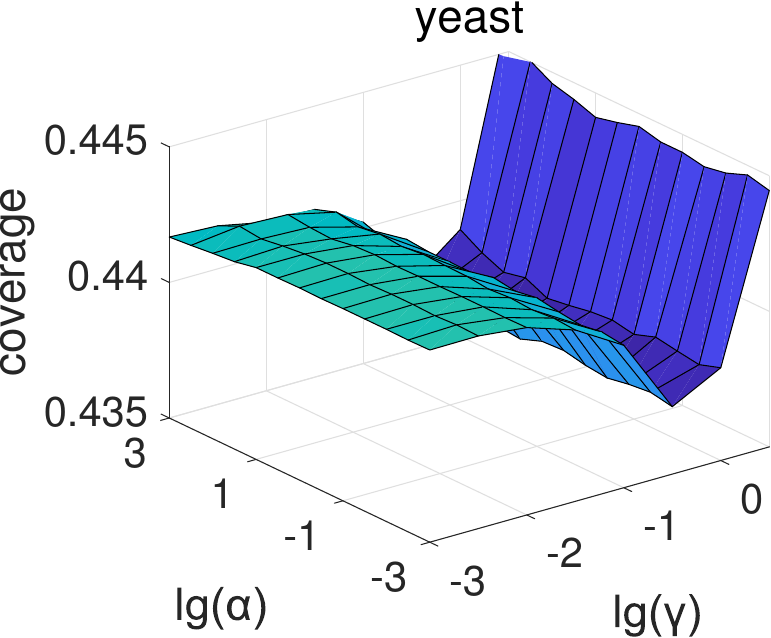}}
	\subfigure[Varying $\alpha$ and $\gamma$ on \textsl{genbase}]{
		\includegraphics[width=0.3\textwidth]{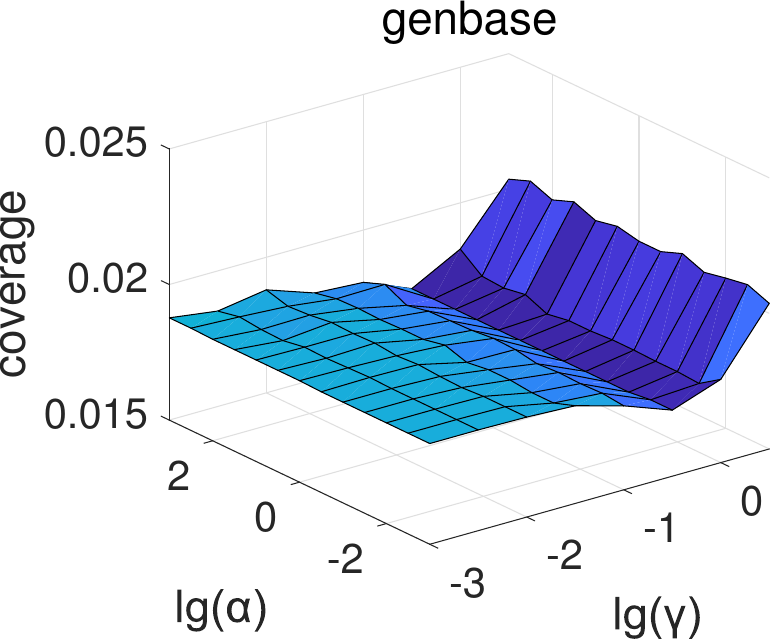}}
	\label{fig:alpha}
	\caption{Influence of the parameters $\alpha$ and $\gamma$ on the \textsl{enron}, \textsl{yeast} and \textsl{genbase} datasets.}
\end{figure}
In summary, IMCC achieves superior performance against other state-of-art multi-label learning algorithms, and the advantage of IMCC is especially pronounced on regular-scale datasets. 

\subsection{Parameter Sensitivity Analysis}	
In this experiment, we study the parameter sensitivity of IMCC on the \textsl{enron} , \textsl{yeast}, and \textsl{genbase} datasets using the $Coverage$ evaluation metric. Concretely, the studied parameters include the regularization parameters $\alpha$, $\beta$, and $\gamma$, and the number of clusters $c$. Note that the importance of learning from virtual examples are controlled by $\alpha$ and $\gamma$, and the importance of the model complexity is controlled by $\beta$. For analyzing the sensitivity of each parameter, we vary one parameter while fixing others at their best setting.
\subsubsection{Influence of Learning from Virtual Examples}

As the importance of learning from virtual examples is controlled by $\alpha$ and $\gamma$, we jointly test the sensitivity of IMCC with respect to $\alpha$ and $\gamma$. The test range of $\alpha$ is $\{10^{-3}, 10^{-2}, \dots, 10^{2}, 10^{3}\}$ and the test range of $\gamma$ is $\{10^{-3},10^{-2},10^{-1},10^0\}$. Figure 2 shows the performance of IMCC on the \textsl{enron}, \textsl{yeast}, and \textsl{genbase} datasets when $\alpha$ and $\gamma$ are varied in the test range. As shown in Figure 2, IMCC is relatively insensitive to the value of $\alpha$. For $\gamma$ that controls the importance of our proposed regularization term, the best performance is achieved at some intermediate value of $\gamma$. Which means, it is helpful to bridge the gap between learning from real examples and virtual examples. Hence the effectiveness of our proposed regularization term is clearly demonstrated. In addition, $10^0$ and $10^{-1}$ are the recommend values of $\alpha$ and $\gamma$, respectively.

\subsubsection{Influence of the Model Complexity}

\begin{figure}[tbp]
	\centering
	\subfigure[Varying $\beta$ on \textsl{enron}]{
		\includegraphics[width=0.3\textwidth]{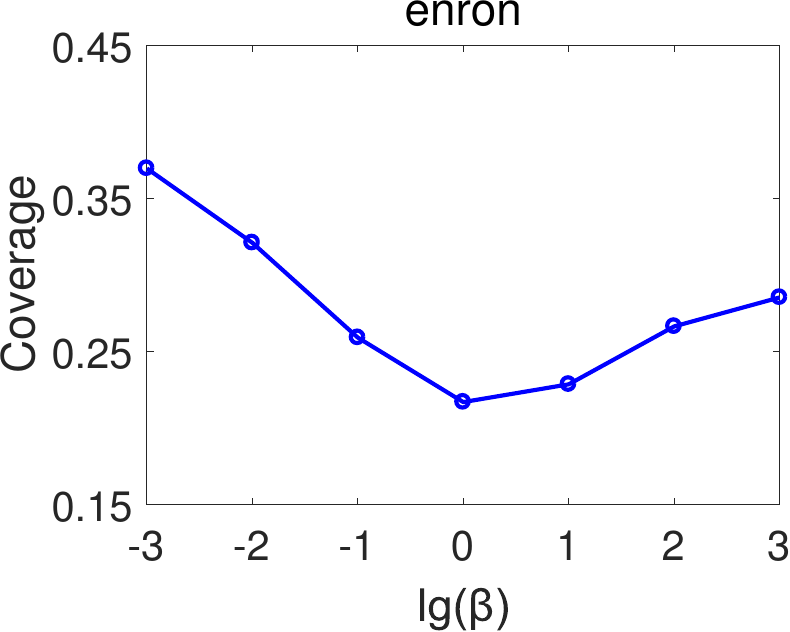}
	}
	\subfigure[Varying $\beta$ on \textsl{yeast}]{
		\includegraphics[width=0.3\textwidth]{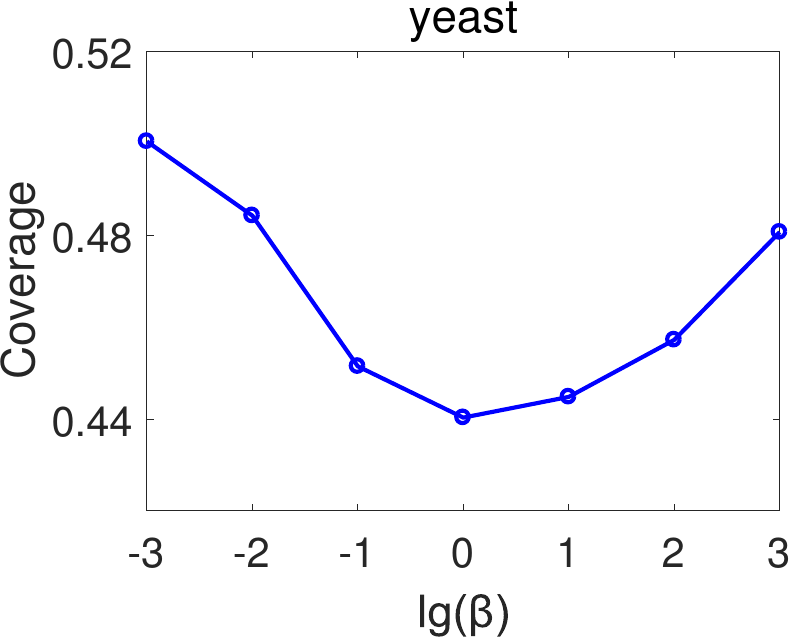}
	}
	\subfigure[Varying $\beta$ on \textsl{genbase}]{
		\includegraphics[width=0.3\textwidth]{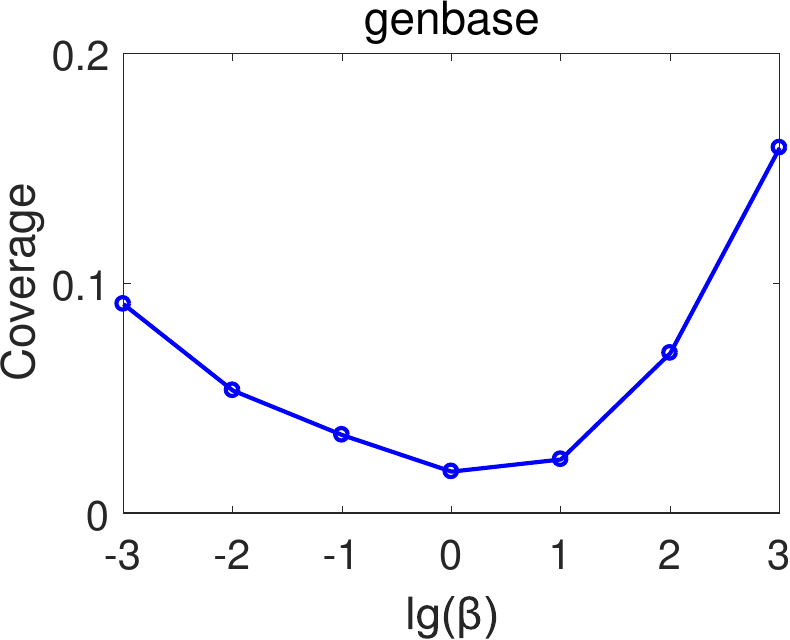}
	}
	\label{fig:beta}
	\caption{Influence of the parameter $\beta$ on the \textsl{enron}, \textsl{yeast} and \textsl{genbase} datasets.}
\end{figure}

The parameter $\beta$ controls the model complexity, and the test value of $\beta$ is chosen from $\{10^{-3},10^{-2},\cdots,10^{2},10^{3}\}$. Figure 3 shows the performance of IMCC on the \textsl{enron}, \textsl{yeast}, and \textsl{genbase} datasets when $\beta$ is varied in the test range. From Figure 3, we can observe that, when $\beta$ is too small, the influence of the term that controls the model complexity will be reduced, which could lead to overfitting. As $\beta$ starts to increase, the performance of IMCC will be improved. However, when $\beta$ is too large, it may overly focus on controlling the model complexity and ignore the importance of model training, which could lead to underfitting, thus the performance of IMCC starts to get worse.
\begin{figure}[htbp]
	\centering	   
	\subfigure[Varying $c$ on \textsl{enron}]{
		\includegraphics[width=0.3\textwidth]{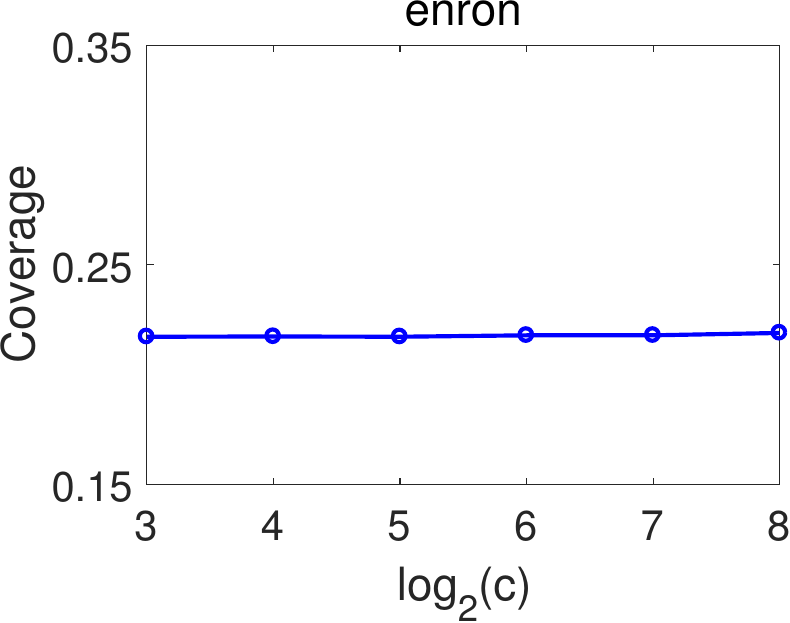}
	}
	\subfigure[Varying $c$ on \textsl{yeast}]{
		\includegraphics[width=0.3\textwidth]{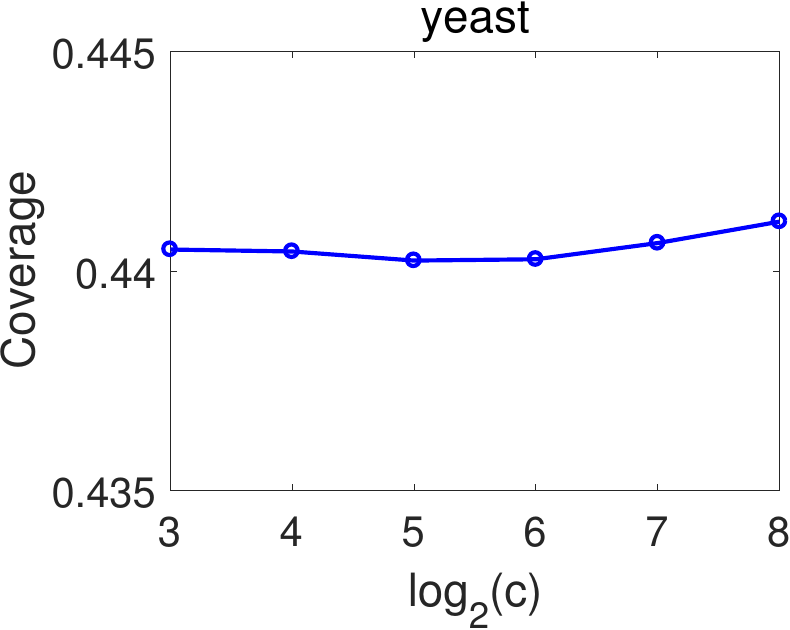}
	}
	\subfigure[Varying $c$ on \textsl{genbase}]{
		\includegraphics[width=0.3\textwidth]{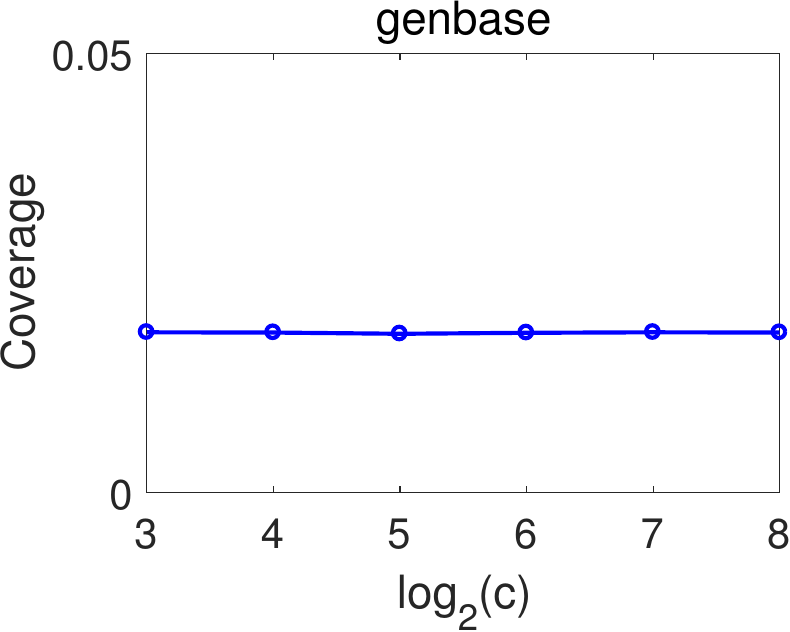}
	}
	\caption{Influence of the parameter $c$ on the \textsl{enron}, \textsl{yeast} and \textsl{genbase} datasets.}
	\label{fig:num}
\end{figure}
Therefore, we can make sure that the value of $\beta$ should not be too large or too small and the best performance is achieved at some intermediate value of $\beta$ such as $\beta=1$.
Such observation exactly agrees with the widely accepted intuition that it is important to balance between overfitting and underfitting.

\subsubsection{Influence of the Number of Clusters}

The parameter $c$ denotes the number of clusters, and the test value of $c$ is chosen from $\{2^{3},...,2^{7},2^{8}\}$. Figure 4 shows the performance of IMCC on the \textsl{enron}, \textsl{yeast}, and \textsl{genbase} datasets when $c$ is varied in the test range. As shown in Figure 4, IMCC achieves rather stable performance when the number of clusters $c$ varies in the test range. Hence IMCC is relatively insensitive to $c$ to some extent. This observation could guide us easily to find a suitable value of the number of clusters.
\section{Conclusion}
In this article, we propose a novel data augmentation approach to enlarge the multi-label training set by generating multiple compact virtual examples from local cluster centers. To the best of our knowledge, this is the first attempt to improve the performance of multi-label learning by data augmentation, since many extensive multi-label learning approaches take into account label correlations explicitly or implicitly to improve the learning performance.
Motivated by the cluster assumption that examples in the same cluster should have the same label, we propose a novel regularization term to bridge the gap between the real examples and virtual examples, which could promote the local smoothness of the learning function. Extensive experimental results demonstrate that our approach outperforms the state-of-the-art counterparts. Specifically, experiments show the average improvement with 7.1\% and 6.0\% over the compared methods by using the \textsl{one error} metric on \textsl{scene} (regular-scale dataset) and \textsl{computer} (large-scale dataset), respectively. These empirical results clearly demonstrate the effectiveness of our proposed approach.

It is worth noting that the clustering technique is used in our proposed approach, and the number of generated virtual examples should be no more than the number of real examples. In the future, we will explore if there exists a better data augmentation approach for multi-label learning without the limitation on the number of generated virtual examples.

\section{Acknowledgement}
This research was partially supported by National Natural Science Foundation of China (Grants No. 61877051), the research program of Chongqing University of Education, China (No. KY2018TZ03), and Natural Science Foundation Project of CQ, China (Grants No. cstc2018jscx-msyb1042 and cstc2017zdcy-zdyf0366).

\bibliography{mybibfile}
\bibliographystyle{ieeetr}
\end{document}